\documentclass[sigconf]{acmart}

\usepackage{enumitem}

\usepackage{multirow}
\usepackage{tabularx}
\usepackage{booktabs}
\usepackage[ruled,linesnumbered]{algorithm2e}
\usepackage{setspace}
\usepackage{array}
\usepackage[normalem]{ulem}
\useunder{\uline}{\ul}{}
\usepackage{graphicx}
\usepackage{subcaption}

\AtBeginDocument{%
  }

\copyrightyear{2025}
\acmYear{2025}
\setcopyright{acmlicensed}\acmConference[KDD '25]{Proceedings of the 31st ACM SIGKDD Conference on Knowledge Discovery and Data Mining V.2}{August 3--7, 2025}{Toronto, ON, Canada}
\acmBooktitle{Proceedings of the 31st ACM SIGKDD Conference on Knowledge Discovery and Data Mining V.2 (KDD '25), August 3--7, 2025, Toronto, ON, Canada}
\acmDOI{10.1145/3711896.3736862}
\acmISBN{979-8-4007-1454-2/2025/08}

\begin{document}

\title{Boosting Bot Detection via Heterophily-Aware Representation Learning and Prototype-Guided Cluster Discovery}

\author{Buyun He}
\affiliation{%
  \institution{University of Science and Technology of China}
  \city{Hefei}
  \country{China}
}
\email{byhe@mail.ustc.edu.cn}

\author{Xiaorui Jiang}
\affiliation{%
  \institution{University of Science and Technology of China}
  \city{Hefei}
  \country{China}
}
\email{xrjiang@mail.ustc.edu.cn}

\author{Qi Wu}
\affiliation{%
  \institution{University of Science and Technology of China}
  \city{Hefei}
  \country{China}
}
\email{qiwu4512@mail.ustc.edu.cn}

\author{Hao Liu}
\affiliation{%
  \institution{University of Science and Technology of China}
  \city{Hefei}
  \country{China}
}
\email{rcdchao@mail.ustc.edu.cn}

\author{Yingguang Yang}
\affiliation{%
  \institution{University of Science and Technology of China}
  \city{Hefei}
  \country{China}
}
\email{dao@mail.ustc.edu.cn}

\author{Yong Liao}
\affiliation{%
  \institution{University of Science and Technology of China}
  \city{Hefei}
  \country{China}
}
\email{yliao@ustc.edu.cn}
\authornote{Corresponding author}

\renewcommand{\shortauthors}{Buyun He et at.}

\begin{abstract}
Detecting social media bots is essential for maintaining the security and trustworthiness of social networks. While contemporary graph-based detection methods demonstrate promising results, their practical application is limited by label reliance and poor generalization capability across diverse communities. Generative Graph Self-Supervised Learning (GSL) presents a promising paradigm to overcome these limitations, yet existing approaches predominantly follow the homophily assumption and fail to capture the global patterns in the graph, which potentially diminishes their effectiveness when facing the challenges of interaction camouflage and distributed deployment in bot detection scenarios. To this end, we propose BotHP, a generative GSL framework tailored to boost graph-based bot detectors through heterophily-aware representation learning and prototype-guided cluster discovery. Specifically, BotHP leverages a dual-encoder architecture, consisting of a graph-aware encoder to capture node commonality and a graph-agnostic encoder to preserve node uniqueness. This enables the simultaneous modeling of both homophily and heterophily, effectively countering the interaction camouflage issue. Additionally, BotHP incorporates a prototype-guided cluster discovery pretext task to model the latent global consistency of bot clusters and identify spatially dispersed yet semantically aligned bot collectives. Extensive experiments on two real-world bot detection benchmarks demonstrate that BotHP consistently boosts graph-based bot detectors, improving detection performance, alleviating label reliance, and enhancing generalization capability.
\end{abstract}



\maketitle

\section{Introduction}

The proliferation of social network bots, automated entities programmatically designed to mimic human behavior~\cite{cresci2020decade,ferrara2022twitter}, has significantly undermined the integrity of online social ecosystems. These sophisticated bots orchestrate coordinated malicious campaigns encompassing misinformation dissemination~\cite{varol2017online,wu2023heterophily}, opinion manipulation~\cite{ferrara2017disinformation,ferrara2020characterizing}, and election interference~\cite{rossi2020detecting,ferrara2016predicting}. In response to these threats, substantial research efforts have been devoted to developing robust bot detection methods to preserve authentic user interactions and information credibility~\cite{yang2023fedack,hays2023simplistic,liu2023botmoe}.

With the advancements in Graph Neural Networks (GNNs), graph-based bot detection methods have garnered significant attention from both academia and industry~\cite{feng2022twibot,shi2023mgtab}. By modeling accounts as nodes and their interactions as edges, these approaches leverage inherent network topology to enhance detection accuracy and achieve state-of-the-art performance~\cite{feng2022twibot}. However, existing graph-based bot detection approaches are predominantly built on supervised learning paradigms~\cite{he2024dynamicity,feng2022heterogeneity,feng2021botrgcn,liu2023botmoe}, which rely on labeled information for supervision from a specific data distribution. Therefore, these methods not only necessitate labor-intensive manual annotations~\cite{shi2023mgtab,feng2021twibot} but also exhibit poor generalization across diverse user communities~\cite{hays2023simplistic,yang2020scalable}, resulting in significant limitations when deployed in real-world applications.

Generative Graph Self-Supervised Learning (GSL) presents a promising paradigm to reduce label dependency and enhance generalization capability by discerning intrinsic patterns from diverse distributions through pretext tasks~\cite{liu2022graph,xie2022self}. While generative GSL has demonstrated remarkable success in various domains~\cite{ye2023graph,yin2024gamc}, its application to bot detection remains largely unexplored. As illustrated in Figure~\ref{fig:intro}, we systematically identify two fundamental challenges when adapting generative GSL to this domain from both individual and cluster perspectives:
\begin{figure}
  \centering
    \includegraphics[width=0.95\linewidth]{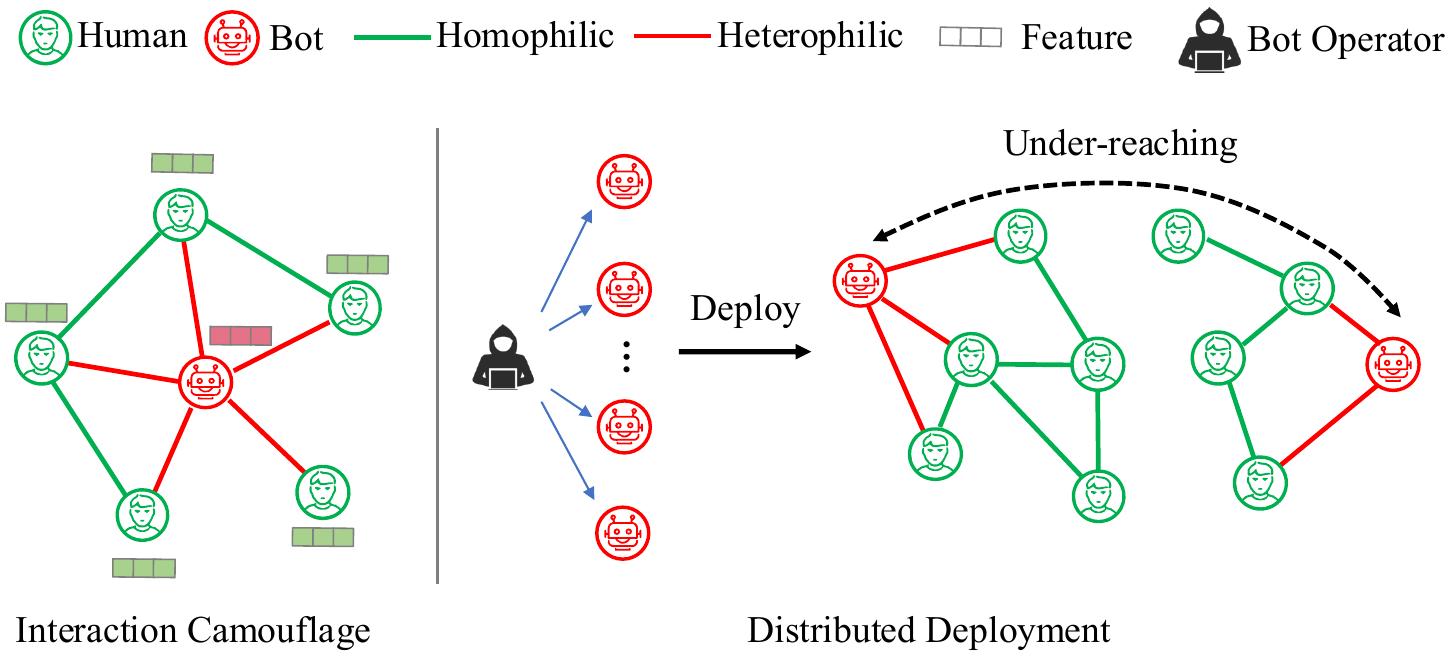}
  \caption{\textit{Interaction Camouflage} and \textit{Distributed Deployment} are two challenges when adapting generative GSL to social network bot detection. Sophisticated bots interact with legitimate users to evade detection, while bot clusters are deployed in a distributed manner to expand influence.}
  \label{fig:intro}
\end{figure}

\begin{itemize}[leftmargin=*]
    \item \textbf{Interaction Camouflage.} From an individual perspective, sophisticated bots strategically interact with legitimate users, resulting in heterophilic connections and distinct characteristics among connected accounts to evade detection~\cite{williams2020homophily,yang2024sebot,liu2023botmoe}. However, contemporary generative GSL methods operate under the implicit homophily assumption~\cite{li2023s}. They employ homophilic encoders that smooth features across connected nodes and leverage pretext tasks, e.g., edge reconstruction, optimized for neighborhood consistency~\cite{hou2022graphmae,tan2022mgae}. Consequently, conventional approaches inherently emphasize node commonality and neglect node uniqueness~\cite{lin2023multi}, making them ineffective in identifying advanced bots that employ interaction camouflage strategies.
    \item \textbf{Distributed Deployment.} From a cluster perspective, bot operators typically deploy a group of collaborative bots, known as a botnet, rather than individual bots to carry out coordinated campaigns~\cite{feng2022twibot}. Furthermore, semantically synchronized bots within a cluster are typically deployed in a distributed manner, which facilitates coordinated campaigns across topologically isolated accounts to expand influence~\cite{cresci2020decade}. While conventional generative GSL methods demonstrate proficiency in capturing \textit{local} patterns via micro-level pretext tasks, such as node-wise feature reconstruction and pair-wise edge reconstruction, they fundamentally lack macro-level pretext tasks to capture \textit{global} information about the graph~\cite{wang2024generative,zheng2024protomgae}. Consequently, these methods struggle to exploit the latent global consistency of bot clusters, hindering their ability to identify topologically dispersed bot collectives.
\end{itemize}

To tackle the aforementioned challenges, we propose \textbf{BotHP}, a generative graph self-supervised learning framework tailored to boost graph-based \textbf{Bot} detectors via \textbf{H}eterophily-aware representation learning and \textbf{P}rototype-guided cluster discovery. BotHP innovatively addresses both interaction camouflage and distributed deployment by refining the encoder architecture and pretext tasks. Specifically, BotHP employs a dual-encoder design consisting of complementary components: a graph-aware encoder that captures node commonality through a message passing mechanism, and a \textit{lightweight} graph-agnostic encoder that preserves node uniqueness via ego feature encoding. This architectural synergy enables simultaneous modeling of homophily and heterophily, effectively countering sophisticated bots' interaction camouflage strategies, without altering the structure of existing graph-based detectors. To identify bot collectives that are spatially dispersed yet semantically aligned, we introduce learnable prototypes to encode latent cluster semantics, coupled with a macro-level cluster discovery pretext task to capture global information about social networks. This mechanism enforces global consistency by enhancing similarity among accounts within the same semantic cluster, thereby revealing potential associations between spatially isolated accounts. Following a widely adopted pre-training and fine-tuning scheme, BotHP first discerns universal patterns within social networks by combining micro-level feature reconstruction and macro-level cluster discovery pretext tasks, and then refines its representations for bot detection by extracting task-specific knowledge. In particular, this paper makes the following contributions:

\begin{itemize}[leftmargin=*]
    \item To the best of our knowledge, we are the first to introduce the application of the generative GSL paradigm to boost existing graph-based bot detectors.
    \item We propose a generative GSL framework specifically designed for bot detection scenarios, addressing the challenges of interaction camouflage and distributed deployment through refinements of encoder architecture and pretext tasks.
    \item We introduce prototype-guided cluster discovery to capture latent global consistency and identify spatially dispersed yet semantically aligned bot collectives.
    \item We conduct extensive experiments on real-world and large-scale bot detection benchmarks, which demonstrate the effectiveness of BotHP in improving detection performance, alleviating label reliance, and enhancing generalization capability.
\end{itemize}

\section{Related Work}
\subsection{Graph-based Social Bot Detection}
Graph-based bot detection methods represent users as nodes and interactions as edges to identify advanced social bots by incorporating the topological structure of social networks~\cite{feng2022twibot}. BotRGCN~\cite{feng2021botrgcn} proposes to model the diverse interactions within social networks via a relational graph convolutional neural network. RGT~\cite{feng2022heterogeneity} further explores the influence heterogeneity through a semantic attention mechanism. To detect the ever-evolving social bots, BotDGT~\cite{he2024dynamicity} incorporates the dynamic nature of social networks. Although these supervised approaches have achieved superior performance, they exhibit a heavy reliance on labeled data and limited generalization across user communities. Several studies introduce GSL for boosting graph-based bot detectors. CBD~\cite{zhou2023detecting} employs contrastive GSL, following the pre-training and fine-tuning paradigm, to combat the issue of label scarcity. SEBot~\cite{yang2024sebot} leverages contrastive learning to capture the consistency between hierarchical information and local information. However, generative GSL remains largely unexplored in the bot detection scenario, a gap we aim to bridge in this paper.

\subsection{Generative Graph Self-supervised Learning}

Generative GSL is a powerful paradigm for reducing label dependency and improving generalization~\cite{xie2022self}, especially advantageous in scenarios where labeled data is scarce, such as recommendation systems~\cite{ye2023graph} and fake news detection~\cite{yin2024gamc}. It generally focuses on reconstructing missing parts of input data within a graph context. GAE~\cite{kipf2016variational} uses a GCN as an encoder and a dot-product decoder to reconstruct the adjacency matrix. MGAE~\cite{tan2022mgae} randomly masks a large proportion of edges and then adopts a cross-correlation decoder to recover them. GraphMAE~\cite{hou2022graphmae} achieves state-of-the-art node classification performance by masking node features and reconstructing them using the features of neighboring nodes. Despite their success, these approaches rely on the homophily assumption~\cite{hou2022graphmae,tan2022mgae,li2023s} and struggle to capture global information~\cite{wang2024generative,zheng2024protomgae}, which diminishes their effectiveness in bot detection scenarios. Building upon prior work, we aim to design a generative GSL framework to counteract the challenges of interaction camouflage and distributed deployment by refining the encoder architecture and pretext tasks.

\subsection{Graph Neural Networks beyond Homophily}
Graph Neural Networks (GNNs) traditionally rely on the homophily assumption, which suggests that nodes with similar features or labels are more likely to be connected~\cite{zheng2022graph,luan2022revisiting,luan2024heterophilic}. However, many real-world graphs exhibit heterophily, where connected nodes may exhibit dissimilar distributions, including social networks~\cite{barranco2019heterophily}. In such cases, homophilic GNNs that primarily reinforce similarities among neighboring nodes~\cite{bo2021beyond,li2022finding} may even perform worse compared to graph-agnostic models like Multilayer Perceptrons (MLPs) that rely exclusively on ego node features~\cite{zhu2020beyond}. Recent works have explored encoding ego features separately from neighboring features to maintain node distinctions, thereby emphasizing node uniqueness. H2GCN~\cite{zhu2020beyond} separates the ego and neighbor embeddings during propagation. MVGE~\cite{lin2023multi} further integrates an MLP, independent of the message-passing process, to ensure distinguishable node representations. Inspired by prior work, BotHP leverages a dual encoder architecture to capture commonality and preserve uniqueness, mitigating the interaction camouflage issue.

\begin{figure*}[t]
    \centering
    \includegraphics[width=0.75\linewidth]{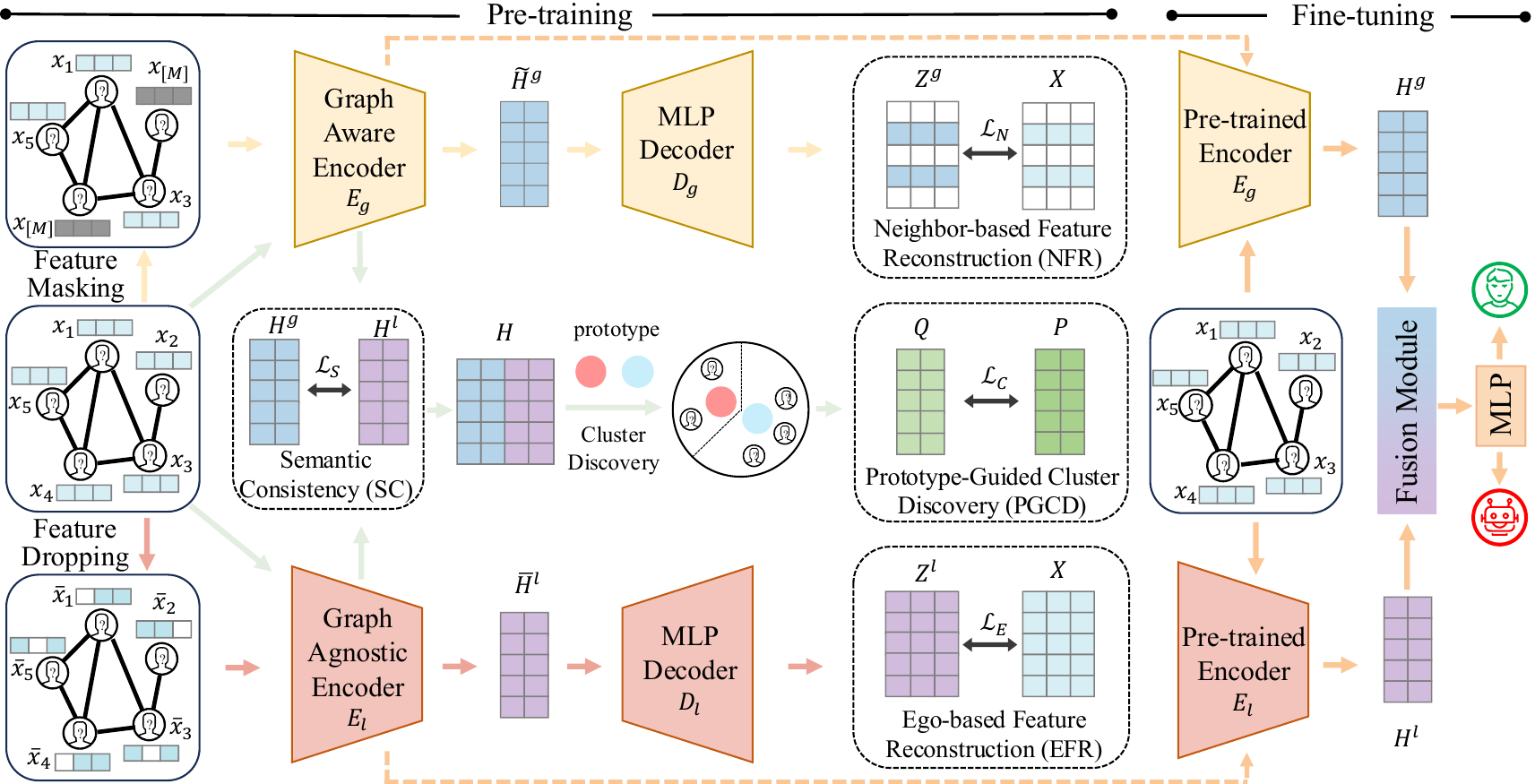}
    \caption{An overview of BotHP, which leverages a dual encoder architecture to address the interaction camouflage issue and employs a prototype-guided cluster discovery mechanism to capture latent global consistency. Following the pre-training and fine-tuning scheme, the dual encoder is first leveraged to discern universal patterns within social networks through multiple pretext tasks and then fine-tuned alongside a classification decoder to extract task-specific knowledge for bot detection.
    }
    \label{fig:framework}
\end{figure*}
\section{Methodology}

Figure~\ref{fig:framework} illustrates BotHP, a generative GSL framework designed to boost graph-based bot detectors. It employs a dual encoder architecture consisting of a graph-aware encoder to capture node commonality and a graph-agnostic encoder to preserve node uniqueness, effectively countering the interaction camouflage, without modifying the structure of existing graph-based detectors. A prototype-guided cluster discovery mechanism is employed to identify distributed bot collectives by learning latent semantic prototypes and enforcing global consistency across topologically dispersed accounts. During the pre-training phase, BotHP first discerns universal patterns through multiple pretext tasks from both micro and macro perspectives. In the fine-tuning phase, the dual encoder is refined alongside a classification decoder to extract task-specific knowledge for bot detection. The pseudocode for BotHP is provided in Appendix~\ref{sec:algorithm}.

\subsection{Dual Encoder Architecture}
In this section, we present the dual encoder architecture, which includes a graph-aware encoder that captures node commonality and a graph-agnostic encoder that preserves node uniqueness. This design enables the simultaneous modeling of both homophily and heterophily, effectively countering interaction camouflage strategies. Notably, BotHP supports various homophilic GNN architectures for the graph-aware encoder, providing flexibility and scalability for real-world applications. Additionally, the graph-agnostic encoder is lightweight, with relatively low computational overhead.

\subsubsection{Graph-aware Encoder}
A neural network that incorporates feature aggregation based on graph structure is referred to as a graph-aware model~\cite{luan2024heterophilic}. Due to the feature aggregation, graph-aware models tend to reinforce similarities among neighboring nodes, capturing node commonality~\cite{bo2021beyond,li2022finding,lin2023multi}. The graph-aware encoder, $E_{g}$, is implemented using a message passing mechanism, formulated as follows:
\begin{equation}
    \begin{aligned}
        m_{j\to i}^{(k)} &= \mathrm{MESSAGE}^{(k)}\Big(h_{i}^{(k-1)}, h_{j}^{(k-1)}\Big) \\
        m_{i}^{(k)} &= \mathrm {AGGREGATE}^{(k)}\Big(m_{j\to i}^{(k)}, \forall j \in \mathcal{N}(i)\Big) \\
        h_{i}^{(k)} &= \mathrm {UPDATE}^{(k)}\Big(h_{i}^{(k-1)}, m_{i}^{(k)}\Big),
    \end{aligned}
\end{equation}

where $h_i^{(k)}$ denotes the hidden state for the user $i$ in layer $k$, $m_{j\to i}^{(k)}$ is the message passed from user $j$ to user $i$, $\mathcal{N}(i)$ denotes the neighboring nodes of user $i$, and $m_{i}^{(k)}$ is the aggregated message for user $i$. MESSAGE, AGGREGATE and UPDATE represent the corresponding operations with learned parameters. The final output of the graph-aware encoder is $h_{i}^{g} = h_{i}^{(K')}$, where $K'$ is the number of message-passing layers.

\subsubsection{Graph-agnostic Encoder}
A neural network that disregards graph structure information is referred to as a graph-agnostic model~\cite{luan2024heterophilic}. Since graph-agnostic models rely solely on the features of individual nodes, they focus on the intrinsic properties of each node without being influenced by potentially dissimilar neighbors, thereby maximizing the preservation of node uniqueness and distinction~\cite{zhu2020beyond,lin2023multi}. The graph-agnostic encoder, $E_{l}$, is implemented using linear transformation operations, defined as follows:
\begin{equation}
    h^{l}_{i} = W_{1} \cdot \sigma(W_{0}\cdot x_{i} + b_{0}) + b_{1},
\end{equation}
where $h$ denotes the hidden state, $\sigma$ is the activation function, $W$ represents the weight matrix, and $b$ denotes the bias term.

\subsection{Universal Pattern Discernment}
To discern universal patterns within social networks across diverse communities without relying on explicit labels, we enable the dual encoder to learn semantic information through pretext tasks ranging from micro-level feature reconstruction and macro-level cluster discovery during the pre-training phase. 

\subsubsection{Neighbor-based Feature Reconstruction}
\label{sec:Neighbor-based Feature Reconstruction}
To enhance the ability of the graph-aware encoder to capture node commonality, we design a neighbor-based feature reconstruction pretext task. In this task, we randomly mask node features and reconstruct them using information derived from neighboring nodes.

Given a social network $\mathcal{G} = (\mathcal{V}, \mathcal{E}, X)$, we first randomly sample a subset of nodes $\mathcal{V}_m \in \mathcal{V}$ for feature masking. For each node in $\mathcal{V}_m$, we replace its original feature with a learnable vector $x_{\left [ M \right ] } \in \mathbb{R}^d $, where $d$ denotes the dimension of the original feature. The masked feature matrix $\widetilde{X}$ is defined as follows:
\begin{equation}
    \widetilde{x}_{i} = \left\{\begin{matrix}
     x_{\left [ M \right ] } & v_{i} \in \mathcal{V}_m\\ 
     x_{i} & v_{i} \notin  \mathcal{V}_m
    \end{matrix}\right.,
    \label{eq:feature_mask}
\end{equation}
where $\widetilde{x}_{i}$ denotes the masked feature of user $i$. Inspired by HGMAE~\cite{tian2023heterogeneous}, we employ a dynamic masking mechanism that gradually increases the feature mask rate based on a linear scheduling function to adaptively adjust the modeling difficulty, forcing the encoder to maintain stable performance despite the varying amount of input information~\cite{gupta2022maskvit}.

Then we forward the masked feature matrix $\widetilde{X}$ and the edge set $\mathcal{E}$ into the graph-aware encoder $E_{g}$:
\begin{equation}
    \widetilde{H}^{g} = E_{g}(\widetilde{X}, \mathcal{E}),
    \label{eq:gnn_encoder}
\end{equation}
where $\widetilde{H}^{g} \in \mathbb{R}^{N \times d_{h}}$ denotes the latent user embeddings generated by the graph-aware encoder with masked features, $N$ is the number of users, and $d_{h}$ denotes the hidden dimension. After obtaining the latent user embeddings $\widetilde{H}^{g}$, we employ an MLP decoder $D_{g}$ to reconstruct the original user features:
\begin{equation}
    Z^{g} = D_{g}(\widetilde{H}^{g}),
    \label{eq:gnn_decoder}
\end{equation}
where $Z^{g} \in \mathbb{R}^{N \times d}$ denotes the reconstructed node features.

The reconstruction error is measured using Mean Squared Error (MSE), which is a standard criterion for evaluating the alignment between reconstructed and original features in generative methods. The neighbor-based feature reconstruction loss is defined as:
\begin{equation}
    \mathcal{L}_{\mathrm{N}}=\frac{1}{|\mathcal{V}_m|} \sum_{v_{i} \in \mathcal{V}_m}(||x_{i} - z^{g}_{i}||_{2}^{2}), 
    \label{eq:L_N}
\end{equation}
where $\mathcal{L}_{\mathrm{N}}$ denotes the reconstruction loss over masked users.

\subsubsection{Ego-based Feature Reconstruction}
To enhance the ability of the graph-agnostic encoder to preserve node uniqueness within social networks, we propose an ego-based feature reconstruction pretext task. In contrast to neighbor-based feature reconstruction, which relies on information from neighboring nodes for feature restoration, the ego-based approach reconstructs a node's feature exclusively based on its intrinsic properties, independent of any neighborhood influence.

To further improve the robustness of the graph-agnostic encoder and mitigate the risk of overfitting to specific patterns, we introduce perturbations via a stochastic dropout mechanism to generate a corrupted feature matrix:
\begin{equation}
    \overline{X} = \mathrm{DROPOUT}(X, p), 
    \label{eq:feature_drop}
\end{equation}
where DROPOUT represents the feature dropping operation, and $p$ is the drop probability. Subsequently, we utilize the graph-agnostic encoder to encode the perturbed feature matrix and adopt an MLP decoder $D_{l}$ to map the latent user embeddings $\overline{H}^{l}$ back to the original input:
\begin{equation}
    \overline{H}^{l} = E_{l}(\overline{X})\quad Z^{l} = D_{l}(\overline{H}^{l}),
    \label{eq:mlp_encoder_and_decoder}
\end{equation}
where $\overline{H}^{l}\in \mathbb{R}^{N \times d_{h}}$ denotes the latent user embeddings generated by the graph-agnostic encoder with corrupted features, and $Z^{l} \in \mathbb{R}^{N \times d}$ denotes the reconstructed features. Notably, $D_{l}$ is shallower than $D_{g}$ because the graph-aware encoder captures higher-order semantic information within the social network, which requires a deeper decoder for accurate reconstruction.

The loss function for ego-based feature reconstruction pretext task is formulated as follows:
\begin{equation}
    \quad\mathcal{L}_{\mathrm{E}}=\frac{1}{|\mathcal{V}|} \sum_{v_{i} \in \mathcal{V}}(||x_{i} - z^{l}_{i}||_{2}^{2}),
    \label{eq:L_E}
\end{equation}
where $\mathcal{L}_{\mathrm{E}}$ denotes the reconstruction loss over all users.

\subsubsection{Semantic Consistency}
The graph-aware encoder emphasizes structural homophily by aggregating information from neighboring nodes, while the graph-agnostic encoder preserves the intrinsic characteristics of nodes. Consequently, the dual encoder captures different patterns of the same user from two distinct perspectives. However, it is crucial to note that, despite the different representations containing specific information within each, they still represent the same user, which intuitively includes consistent information. Motivated by canonical correlation analysis~\cite{zhang2021canonical}, we introduce the following objective to enhance semantic consistency between the obtained representations:
\begin{equation}
\mathcal{L}_{\mathrm{S}} = \underbrace{\left\| H^{g} - H^{l} \right\|_F^2}_{\text{Invariance}} +  \underbrace{\left\| {H^{g}}^\top H^{g} - I \right\|_F^2 + \left\| {H^{l}}^\top H^{l} - I \right\|_F^2}_{\text{Decorrelation}},
\label{eq:L_S}
\end{equation}
where $H^{g} \in \mathbb{R}^{N \times d_h}$ and $H^{l} \in \mathbb{R}^{N \times d_h}$ denote the outputs of the dual encoder without masking or perturbation, and $\mathcal{L}_{\mathrm{S}}$ represents the semantic consistency loss. In the objective, the invariance term encourages $H^{g}$ and $H^{l}$ to align with each other, thereby converging to consistency. Meanwhile, the decorrelation term enforces the different dimensions of $H^{g}$ and $H^{l}$ to uniformly distribute over the latent space, thus eliminating redundant correlations and preventing model collapse.

\subsubsection{Prototype-Guided Cluster Discovery}
To counteract the challenge of distributed bot deployment and uncover latent global consistency among topologically dispersed yet semantically aligned bot collectives, we design a prototype-guided cluster discovery pretext task based on deep clustering technique~\cite{Xie_Girshick_Farhadi_2016,jiang2024heterogeneity}. This component introduces learnable semantic prototypes to establish cluster centroids in the representation space, coupled with a macro-level pretext task that enforces global consistency among spatially isolated accounts, effectively bridging local topological patterns with global semantic relationships.

Specifically, we first introduce a set of trainable prototypes $C = [c_1, c_2, \dots, c_K] \in \mathbb{R}^{K \times 2d_{h}}$, where $K$ denotes the number of the prototypes, and each prototype $c_i$ serves as an adaptive semantic center of a potential account cluster. We employ K-means clustering~\cite{hartigan1979k} on the output of the dual encoder and initialize the prototypes with the obtained centroids:
\begin{equation}
\min _{\mathbf{c}_{1}, \mathbf{c}_{2}, \ldots, \mathbf{c}_{K}} \sum_{i=1}^{N} \sum_{j=1}^{K}\left\|h_{i}-{c}_{j}\right\|^{2},
\end{equation}
where the concatenated representation $h_{i} = [h^{g}_{i} || h^{l}_{i}]$ combines the outputs of the dual encoder, thereby enhancing the expressiveness of the learned representation.

To address the non-differentiable nature of hard cluster assignments, we measure user-prototype similarity using a Student's t-distribution kernel~\cite{Maaten_Hinton_2008}. This produces soft cluster assignments that reveal latent semantic relationships:
\begin{equation}
q_{ij} = \frac{(1 + \| h_{i} - c_j \|^2 / \alpha)^{-(\alpha + 1) / 2}}{\sum_{k=1}^K (1 + \| h_{i} - c_k \|^2 / \alpha)^{-(\alpha + 1) / 2}},
\label{eq:Q}
\end{equation}
where $\alpha$ is the degree of freedom of the Student’s t-distribution. The probabilistic assignment $q_{ij}$ quantifies how likely user $i$ belongs to the semantic cluster represented by prototype $j$.

Upon acquiring the soft alignment distribution $Q$, we then amplify high-confidence assignments to generate a target distribution $P$ that promotes intra-cluster cohesion and reinforces the spatially dispersed yet semantically
aligned users to converge around the same prototype. The target distribution $p_{ij}$ is computed as:
\begin{equation}
p_{i j} = \frac{q_{i j}^{2} / f_{j}}{\sum_{j^{\prime}} q_{i j^{\prime}}^{2} / f_{j^{\prime}}},
\label{eq:P}
\end{equation}
where $f_{j} = {\textstyle \sum_{i}^{}} q_{ij}$ denotes soft cluster frequencies. This formulation squares and normalizes the assignments in $Q$, effectively amplifying confidence levels. The optimization objective of the prototype-guided cluster discovery pretext task is defined as the Kullback-Leibler (KL) divergence between $P$ and $Q$:
\begin{equation}
 \mathcal{L}_{\mathrm{C}} = K L(P \| Q) = \sum_{i} \sum_{j} p_{i j} \log \frac{p_{i j}}{q_{i j}}.
 \label{eq:L_D}
\end{equation}
By minimizing this loss, the dual encoder iteratively aligns user representations with their corresponding semantic prototypes, enhancing similarity among
users within the same semantic cluster regardless of their topological positions. As a result, this mechanism reveals potential associations between spatially isolated users and discovers latent bot collectives despite topological dispersion. Notably, the target distribution $P$ is updated every $T$ epochs to ensure efficient model training and mitigate potential instability.

\subsubsection{Model Pre-training}
The overall objective to be minimized is defined as a weighted sum of the previously mentioned losses:
\begin{equation}
    \mathcal{L}_{P} = \mathcal{L}_{\mathrm{N}} + \mathcal{L}_{\mathrm{E}} + \mathcal{L}_{\mathrm{S}} + \mathcal{L}_{\mathrm{C}}.
    \label{eq:pretrain_loss}
\end{equation}
By jointly optimizing these losses, BotHP enables the dual encoder to capture universal patterns within social networks from micro and macro perspectives, resulting in more comprehensive user embeddings for bot detection.

\subsection{Task-Specific Knowledge Extraction}
To extract task-specific knowledge for bot detection, we fine-tune the dual encoder in conjunction with a classification decoder, incorporating an embedding fusion module to integrate the embeddings from the dual encoder for robust user representations.

\subsubsection{Embedding Fusion Module}
\label{sec:embedding_fusion_module}
The graph-aware encoder and graph-agnostic encoder are employed to capture node commonality and uniqueness, respectively, generating user embeddings from different semantic spaces. To produce more robust user representations for bot detection with varying levels of homophily, we introduce an Embedding Fusion Module (EFM) with an attention mechanism to adaptively determine the importance of each embedding. The attention mechanism in the fusion module is formulated as follows:
\begin{equation}
    s^{*}_{i} = q^\top \mathrm{Tanh}(W·h^{*}_{i} + b),  \alpha^{*}_i = \frac{\exp(s^{*}_{i})}{ \exp(s^{g}_{i}) + \exp(s^{l}_{i})}^{}, u_{i} = \sum_{\ast  \in \left \{ g, l \right \} }^{}  \alpha^{*}_i · h^{*}_{i},
\end{equation}
where $* \in \{g, l\}$, $q$ denotes a learnable attention vector, $s^{*}$ is the importance score, $\alpha^{*}$ denotes the attention weights, and $u$ denotes the final user representation. Leveraging the EFM module, we form a unified yet discriminative representation space for bot detection.

\subsubsection{Model Fine-tuning}
A downstream decoder $D_{d}$, composed of the proposed fusion module and a linear transform module, is utilized to obtain the bot detection result. We first generate the user representations with the embedding fusion module:
\begin{equation}
    H^{g} = E_{g}(X, \mathcal{E}), \quad H^{l} = E_{l}(X),
    \quad U = \mathrm{EFM}(H^{g}, H^{l}),
    \label{eq:user_rep}
\end{equation}
where $U \in \mathbb{R}^{N \times d_{h}}$ denotes the user representations. Then the user representations are passed into a linear layer to predict user labels. A binary cross-entropy loss is employed to optimize the model with an $L_{2}$ regularization term to combat overfitting. The process is formulated as follows: 
\begin{equation}
    \hat{Y} = \mathrm{MLP}(U),
    \quad \mathcal{L}_{F} = \sum_{i=1}^{N}  {y}_{i}\log(\hat{y}_{i})  + \lambda \sum_{w\in\theta }^{} w^{2},
    \label{eq:finetune_loss}
\end{equation}
where $\hat{y}_{i}$ is the predicted output, $y_{i}$ is the corresponding ground truth label, and $\theta$ denotes all trainable model parameters, including those of $E_{g}$, $E_{l}$, and $D_{d}$.
\section{Experiments}
In this section, we conduct empirical evaluations of BotHP on benchmark datasets to investigate the following research questions:
\begin{itemize}[leftmargin=*]
  \item \textbf{RQ1:} How does BotHP perform compared to other representative baseline methods? Can it improve the detection performance of graph-based bot detectors?
  \item \textbf{RQ2:} What are the effects of different components in BotHP?
  \item \textbf{RQ3}: How does BotHP perform concerning different hyperparameters?
  \item \textbf{RQ4:} Can BotHP alleviate label reliance? 
  \item \textbf{RQ5:} Can BotHP enhance the generalization capability? 
\end{itemize}

\subsection{Experimental Setup}
\subsubsection{Dataset}
We conduct experiments on two real-world social bot detection datasets: TwiBot-20~\cite{feng2021twibot} and MGTAB~\cite{shi2023mgtab}. These datasets encompass a diverse range of entities and relationships, facilitating the evaluation of graph-based bot methods. We represent users as nodes and construct an interaction graph based on following and follower relationships. The details of both datasets are presented in Table~\ref{tab:dataset}, where \textbf{homo} indicates the edge homophily ratio, i.e., the proportion of edges connecting nodes with the same class label, as defined by~\cite{zhu2020beyond}. For TwiBot-20, we follow the official dataset splits provided in the benchmark. For MGTAB, which does not include official splits, we create a 7:2:1 random split for the training, validation, and test sets to ensure a fair comparison.

\subsubsection{Baselines}
Our evaluation of BotHP involves a systematic comparison against a range of established baseline methods. These baselines are organized into five primary categories: homophilic methods, heterophilic methods, supervised bot detection techniques, contrastive GSL approaches specifically adapted for bot detection, and generative GSL methods. Detailed descriptions of these baselines are provided below:

\begin{itemize}[leftmargin=*]
\item \textbf{GCN~\cite{kipf2016semi}} is a classic homophilic GNN that aggregates information from neighbors to learn node representations.
\item \textbf{GAT~\cite{velickovic2017graph}} employs an attention mechanism to weigh the influence of neighboring nodes.
\item \textbf{H2GCN~\cite{zhu2020beyond}} leverages the separation of ego and neighbor embeddings, higher-order neighborhoods, and intermediate representation combinations to enhance heterophilic graph learning.
\item \textbf{FAGCN~\cite{bo2021beyond}} defines enhanced low-pass and high-pass filters and employs a self-gating mechanism to adaptively integrate low-frequency signals, high-frequency signals, and raw features during message passing.
\item \textbf{RGCN~\cite{feng2021botrgcn}} applies relational graph convolutional networks to learn user representations.
\item \textbf{RGT~\cite{feng2022heterogeneity}} employs relational graph transformers to model the heterogeneous relationships and influence in the Twitter space.
\item \textbf{BotMoE~\cite{liu2023botmoe}} uses a community-aware mixture of experts architecture to integrate multimodal information, enhancing detection performance and generalization.
\item \textbf{BotDGT~\cite{he2024dynamicity}} leverages a dynamic graph transformer to capture the dynamic nature of social networks.
\item \textbf{CBD~\cite{zhou2023detecting}} is a contrastive GSL framework designed for bot detection, which pre-trains a graph-aware encoder with contrastive learning and then fine-tunes it with a consistency loss.
\item \textbf{SEBot~\cite{yang2024sebot}} incorporates hierarchical information, modifies RGCN for heterophily, and employs contrastive learning to improve semantic consistency.
\item \textbf{GAE~\cite{kipf2016variational}} utilizes an autoencoder to reconstruct the adjacency matrix for learning node embeddings.
\item \textbf{MGAE~\cite{tan2022mgae}} masks a proportion of edges and then adopts a cross-correlation decoder to recover them.
\item \textbf{GraphMAE~\cite{hou2022graphmae}} introduces node feature masking and remasking strategies, with GNNs as the encoder and decoder to reconstruct masked node features.
\end{itemize}

\begin{table}[t]
\centering
\caption{Statistics of TwiBot-20 and MGTAB.}
\label{tab:dataset}
\resizebox{0.45\textwidth}{!}{%
\begin{tabular}{cccccc}
\hline
\toprule
\textbf{Dataset}                    & \textbf{\#users}                & \textbf{\#interactions}           & \textbf{class} & \textbf{\#class} & \textbf{homo}                     \\
\midrule
\multirow{2}{*}{Twibot-20} & \multirow{2}{*}{229,580} & \multirow{2}{*}{227,979}   & human & 5,237    & \multirow{2}{*}{0.53} \\
                           &                        &                          & bot   & 6,589    &                          \\
\specialrule{0em}{1pt}{1pt}
\hline
\specialrule{0em}{1pt}{1pt}
\multirow{2}{*}{MGTAB}     & \multirow{2}{*}{10,199} & \multirow{2}{*}{720,695} & human & 7,451    & \multirow{2}{*}{0.92} \\
                           &                        &                          & bot   & 2,748    &       \\      
\bottomrule
\end{tabular}
}
\end{table}

\subsubsection{Implementation}
We implement BotHP using PyTorch~\cite{paszke2017automatic} and PyTorch Geometric~\cite{fey2019fast}. The initial user features are processed using the encoding procedure outlined in recent studies~\cite{feng2022heterogeneity,shi2023mgtab} to facilitate fair comparison. Depending on the choice of graph-aware encoder, we implement two variants: BotHP (RGCN) and BotHP (RGT), which denote the BotHP framework with RGCN~\cite{feng2021botrgcn} and RGT~\cite{feng2022heterogeneity} as the graph-aware encoder, respectively. For all generative GSL methods, we apply the pre-training and fine-tuning scheme and use RGT~\cite{feng2022heterogeneity} as the backbone encoder due to its superior performance on the bot detection task. Given the relatively sparse edges in TwiBot-20 for neighbor-based feature reconstruction, we apply a graph augmentation strategy that adds edges via 2-hop neighbors. The complete code and hyperparameter settings are released on Github\footnote{\url{https://github.com/Peien429/BotHP}}.

\begin{table*}[ht]
\caption{Performance comparison on the TwiBot-20 and MGTAB datasets. We run each method five times and report the mean value along with the standard deviation. The best and second-best results are highlighted with \textbf{bold} and \underline{underline}, respectively. OOM indicates Out-Of-Memory on a 24GB GPU, and \mbox{-} denotes that the method is not scalable to the dataset. We report the relative improvement achieved by the BotHP framework over its graph-aware backbone encoder.
}
\centering
\label{tab:performance}
\scalebox{0.92}{
\begin{tabular}{cc|cccc|cccc}
\hline
\toprule
\specialrule{0em}{1pt}{1pt}
 \multicolumn{2}{c}{\multirow{2}{*}{\textbf{Methods}}} &\multicolumn{4}{|c}{\textbf{TwiBot-20}} &\multicolumn{4}{|c}{\textbf{MGTAB}} \\
\specialrule{0em}{1pt}{1pt}
\cline{3-9}\cline{7-10}
\specialrule{0em}{1pt}{1pt}
\multicolumn{2}{c|}{} &\textbf{Accuracy} &\textbf{F1-score} &\textbf{Precision}  &\textbf{Recall} &\textbf{Accuracy} &\textbf{F1-score} &\textbf{Precision}  &\textbf{Recall} \\
\specialrule{0em}{1pt}{1pt}
\hline
\specialrule{0em}{1pt}{1pt}

\multirow{2}{*}{\shortstack{Homophilic}}
& GCN 
& 83.45$\pm$0.61 
& 85.75$\pm$0.58
& 80.25$\pm$0.37 
& 92.06$\pm$1.07 
& 88.03$\pm$1.03 
& 75.01$\pm$2.08
& 85.05$\pm$2.76 
& 67.11$\pm$1.95 
\\
&GAT 
& 84.97$\pm$0.25 
& 86.96$\pm$0.22
& 81.95$\pm$0.36 
& 92.62$\pm$0.51 
& 89.15$\pm$0.17 
& 79.32$\pm$0.18 
& 81.06$\pm$1.52 
& 77.77$\pm$1.61 
\\
\specialrule{0em}{1pt}{1pt}
\hline
\specialrule{0em}{1pt}{1pt}

\multirow{2}{*}{\shortstack{Heterophilic}}
& FAGCN 
& 85.80$\pm$0.29 
& 87.57$\pm$0.24 
& 83.18$\pm$0.36 
& 92.44$\pm$0.25
& 89.39$\pm$0.69 
& 79.31$\pm$1.67 
& 82.97$\pm$2.41
& 76.15$\pm$3.83 
\\
& H2GCN 
& 86.46$\pm$0.39
& 88.01$\pm$0.40
& 84.46$\pm$0.28
& 91.88$\pm$0.87
& 89.27$\pm$0.22
& 80.07$\pm$0.36
& 78.74$\pm$0.75
& 81.47$\pm$0.76 
\\
\specialrule{0em}{1pt}{1pt}
\hline
\specialrule{0em}{1pt}{1pt}

\multirow{4}{*}{\shortstack{Supervised\\Bot\\Detectors}}
& RGCN 
& 85.71$\pm$0.29
& 87.45$\pm$0.26
& 83.14$\pm$0.33 
& 92.28$\pm$0.56
& 89.50$\pm$0.42 
& 79.43$\pm$0.58
& 83.61$\pm$2.23
& 75.71$\pm$1.60 
\\
& RGT  
& 86.51$\pm$0.11 
& 87.91$\pm$0.14  
& 84.73$\pm$0.45 
& 91.34$\pm$0.47 
& 89.84$\pm$0.66
& 77.77$\pm$1.93 
& \textbf{88.51$\pm$0.72}
& 69.45$\pm$3.36
\\
& BotMoE  
& 86.73$\pm$0.34
& 88.27$\pm$0.30
& 84.17$\pm$0.51
& 92.91$\pm$0.62 
& OOM
& OOM 
& OOM
& OOM
\\
& BotDGT  
& 87.17$\pm$0.22
& \underline{88.76$\pm$0.23}
& 84.60$\pm$0.26
& 93.37$\pm$1.42 
& \mbox{-}
& \mbox{-} 
& \mbox{-}
& \mbox{-} 
\\
\specialrule{0em}{1pt}{1pt}
\hline
\specialrule{0em}{1pt}{1pt}

\multirow{2}{*}{\shortstack{Contrastive\\ GSL}}
& CBD  
& 85.65$\pm$0.32
& 87.39$\pm$0.25
& 82.64$\pm$0.38 
& \underline{93.40$\pm$0.17}
& 90.35$\pm$0.43  
& 81.31$\pm$0.61
& 83.70$\pm$0.85
& 79.08$\pm$1.65 
\\
& SEBot
& \underline{87.23$\pm$0.31}
& 88.71$\pm$0.21  
& \underline{84.93$\pm$0.49}
& 92.74$\pm$0.67 
& 90.54$\pm$0.27
& 82.45$\pm$0.17
& 81.52$\pm$0.45
& \textbf{83.63$\pm$1.23}
\\
\specialrule{0em}{1pt}{1pt}
\hline
\specialrule{0em}{1pt}{1pt}

\multirow{3}{*}{\shortstack{Generative\\GSL}}
& GAE 
& 86.05$\pm$0.15
& 87.35$\pm$0.17
& 84.55$\pm$0.27 
& 90.81$\pm$0.55
& 89.93$\pm$0.74
& 80.72$\pm$1.04
& 79.77$\pm$2.78
& 81.34$\pm$2.37 
\\
& MGAE  
& 86.03$\pm$0.59
& 87.57$\pm$0.54
& 84.26$\pm$0.60 
& 91.16$\pm$0.84 
& 90.28$\pm$0.27  
& 81.65$\pm$0.40
& 83.38$\pm$1.59
& 81.03$\pm$1.92 
\\
& GraphMAE  
& 86.48$\pm$0.27
& 88.02$\pm$0.23
& 84.52$\pm$0.37
& 91.81$\pm$0.35 
& 90.34$\pm$0.78
& 81.89$\pm$1.45
& 82.91$\pm$2.31
& 81.50$\pm$0.74  
\\

\specialrule{0em}{1pt}{1pt}
\hline
\specialrule{0em}{1pt}{1pt}

\multirow{4}{*}{\shortstack{ours}}
& BotHP (RGCN)
& 87.05$\pm$0.16 
& 88.60$\pm$0.19 
& 84.56$\pm$0.34 
& 93.06$\pm$0.55  
& \underline{91.03$\pm$0.48}
& \underline{82.49$\pm$1.03}
& \underline{86.32$\pm$0.93}
& 79.01$\pm$1.62
\\
& Gain
& $\uparrow$ 1.34
& $\uparrow$ 1.15 
& $\uparrow$ 1.42
& $\uparrow$ 0.78  
& $\uparrow$ 1.53
& $\uparrow$ 3.06
& $\uparrow$ 2.71
& $\uparrow$ 3.30
\\

\specialrule{0em}{1pt}{1pt}
\cline{3-10}
\specialrule{0em}{1pt}{1pt}

& BotHP (RGT)
& \textbf{87.73$\pm$0.15}
& \textbf{89.22$\pm$0.11}
& \textbf{85.02$\pm$0.29}
& \textbf{93.84$\pm$0.21} 
& \textbf{91.52$\pm$0.16}
& \textbf{83.74$\pm$0.29}
& 85.99$\pm$0.52
& \underline{81.61$\pm$0.38 }
\\
& Gain
& $\uparrow$ 1.23 
& $\uparrow$ 1.31 
& $\uparrow$ 0.29 
& $\uparrow$ 2.50  
& $\uparrow$ 1.68
& $\uparrow$ 5.97
& $\downarrow$ 2.52
& $\uparrow$ 12.16
\\
\specialrule{0em}{1pt}{1pt}
\bottomrule
\end{tabular}
}
\end{table*}

\subsection{Framework Performance (RQ1)}
\label{sec:Framework Performance}
To address RQ1, we evaluate BotHP alongside several representative baseline methods on two social network bot detection benchmarks and present the results in Table~\ref{tab:performance}, which demonstrate that:

\begin{itemize}[leftmargin=*]
\item BotHP achieves superior bot detection performance. Specifically, BotHP (RGT) consistently outperforms all baseline methods in terms of Accuracy and F1-score on both datasets, while BotHP (RGCN) demonstrates leading performance on MGTAB. These results highlight the effectiveness of the proposed BotHP for bot detection.
\item 
Both variants of BotHP consistently outperform their respective graph-aware backbone encoders in detection performance. This superior performance can be attributed to BotHP's dual-encoder architecture, which incorporates a graph-agnostic encoder that preserves crucial node uniqueness, as well as to its multiple pretext tasks that capture universal social network patterns. Consequently, BotHP significantly enhances the detection performance of graph-based bot detectors.
\item Heterophilic GNNs generally exhibit superior detection performance compared to homophilic GNNs, emphasizing the importance of analyzing heterophily within social networks to counteract interaction camouflage and achieve robust bot detection.
\item Generative GSL methods tend to boost graph-based detectors on MGTAB but degrade performance on TwiBot-20. This inconsistency arises because these methods rely on the homophily assumption, which may introduce an inappropriate inductive bias for bot detection tasks in some cases. In contrast, BotHP leverages the dual encoder architecture to handle varying degrees of homophily and heterophily, providing a robust boost to graph-based detectors.
\end{itemize}

\subsection{Ablation Study (RQ2)}
\label{sec:ablation}
We conduct ablation experiments to thoroughly elucidate the role of the dual encoder architecture and diverse pretext tasks in our proposed BotHP framework. The results of the accuracy from the ablation experiments are presented in Table~\ref{tab:ablation}.

\begin{table}[t]
\caption{Ablation study of the dual encoder architecture and the diverse pretext tasks on Twibot-20 and MGTAB.}
\label{tab:ablation}
\centering
\scalebox{0.95}{
\begin{tabular}{c|c|cc}
\hline
\toprule
\specialrule{0em}{1pt}{1pt}
\textbf{Category}                      & \textbf{Ablation Settings} & \textbf{Twibot-20} & \textbf{MGTAB} \\
\specialrule{0em}{1pt}{1pt}
\hline
\specialrule{0em}{1pt}{1pt}
full model & BotHP (RGT) & \textbf{87.73$\pm$0.15} & \textbf{91.52$\pm$0.16} \\
\specialrule{0em}{1pt}{1pt}
\hline
\specialrule{0em}{1pt}{1pt}
\multirow{2}{*}{encoder type} 
& w/o graph-aware
& 85.59$\pm$0.15     
& 87.26$\pm$0.25 
\\
& w/o graph-agnostic
& 86.84$\pm$0.29  
& 90.69$\pm$0.18 
\\
\specialrule{0em}{1pt}{1pt}
\hline
\specialrule{0em}{1pt}{1pt}
\multirow{4}{*}{pretext task} 
& w/o NFR           
& 87.24$\pm$0.26     
& 90.17$\pm$0.22 
 \\
 & w/o EFR        
 & 87.35$\pm$0.33 
 & 90.75$\pm$0.22 
 \\
 & w/o SC
 & 87.18$\pm$0.24     
 & 90.41$\pm$0.27 
 \\
 & w/o PGCD           
 & 87.10$\pm$0.13
 & 90.85$\pm$0.26 
 \\
\specialrule{0em}{1pt}{1pt}
\bottomrule
\end{tabular}
}
\end{table}

\textbf{Effect of dual encoder architecture.} To investigate the roles of the graph-aware encoder and the graph-agnostic encoder, we remove one encoder and its related pretext tasks, relying solely on the other encoder for bot detection. The removal of the graph-aware encoder led to a significant drop in performance on both datasets, emphasizing the importance of modeling the topological structure of social networks for effective bot detection. When the graph-agnostic encoder is removed, the performance decline is more pronounced on TwiBot-20, while the reduction on MGTAB is less substantial. This result suggests that the graph-agnostic encoder is particularly critical for heterophilic social networks in effectively preserving node uniqueness, thereby ensuring distinguishable user representations. Overall, the dual encoder architecture achieves robust performance across social networks with varying degrees of homophily and heterophily, demonstrating their complementary roles in adversarial social bot detection. A further analysis of the dual encoder architecture is presented in Appendix~\ref{sec:Impact of Dual Encoder Architecture}.

\textbf{Effect of diverse pretext tasks.} BotHP incorporates multiple pretext tasks to discern universal patterns within social networks. We compare the full BotHP framework with variants that omit specific pretext tasks to assess their contributions. The variants w/o NFR and w/o EFR exhibit a notable decline in performance on both datasets,  highlighting the crucial role of micro-level feature reconstruction pretext tasks for bot detection. Additionally, excluding the semantic consistency loss results in decreased performance, underscoring the importance of enforcing alignment and decorrelation between the dual encoder representations to maintain both consistency and diversity in the learned features. Finally, the performance decline in the variant w/o PGCD demonstrates the importance of the macro-level cluster discovery pretext task in capturing the latent global consistency among similar accounts. Overall, the diverse pretext tasks enable the dual encoder to extract valuable knowledge from social networks, leading to a more effective bot detector.

\begin{figure}[t]
	\centering
	\begin{minipage}[c]{0.475\linewidth}
		\centering
		\includegraphics[width=\linewidth]{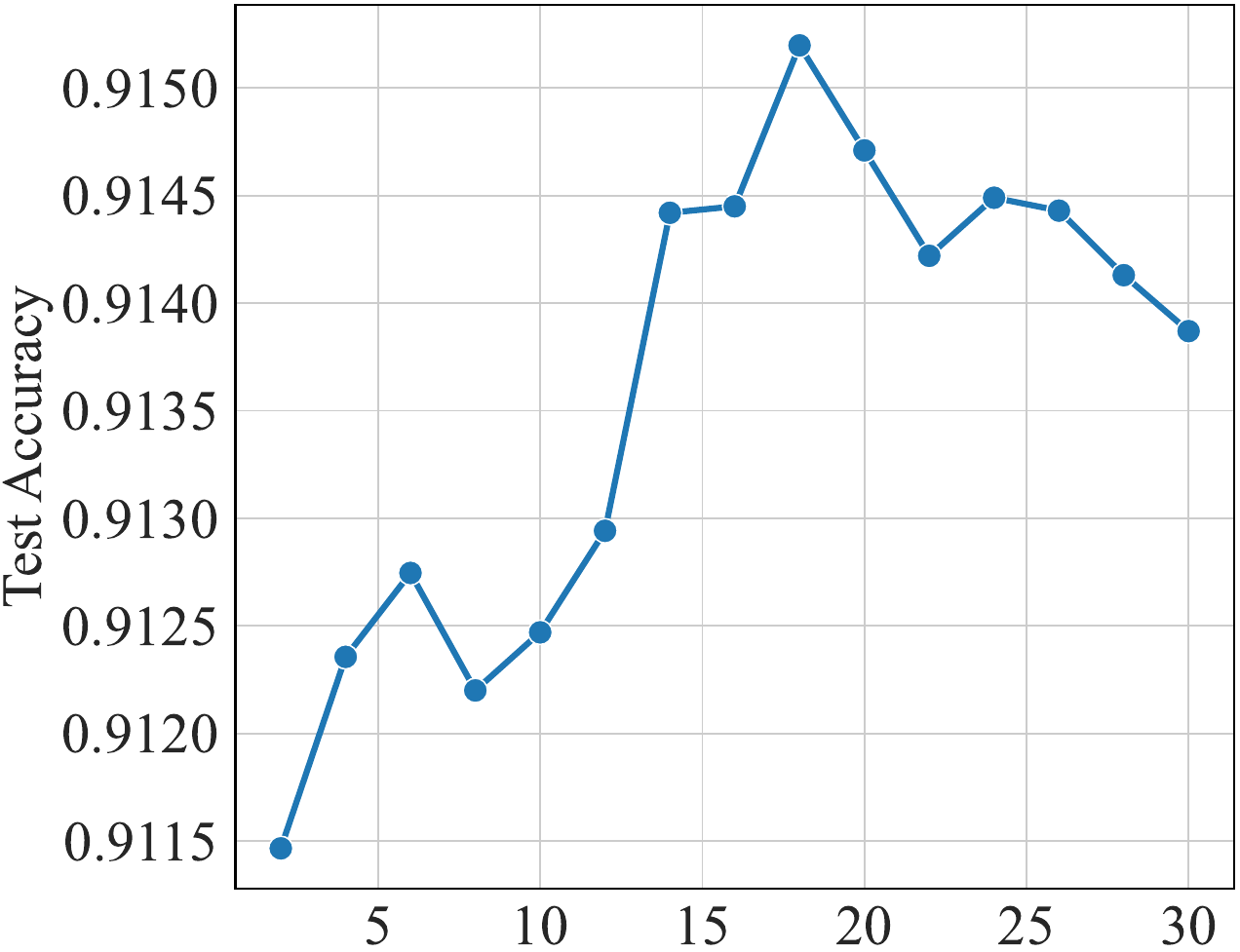}
		\subcaption{Varied number of prototypes}
		\label{fig:cluster_num}
	\end{minipage} 
	\begin{minipage}[c]{0.475\linewidth}
		\centering
		\includegraphics[width=\linewidth]{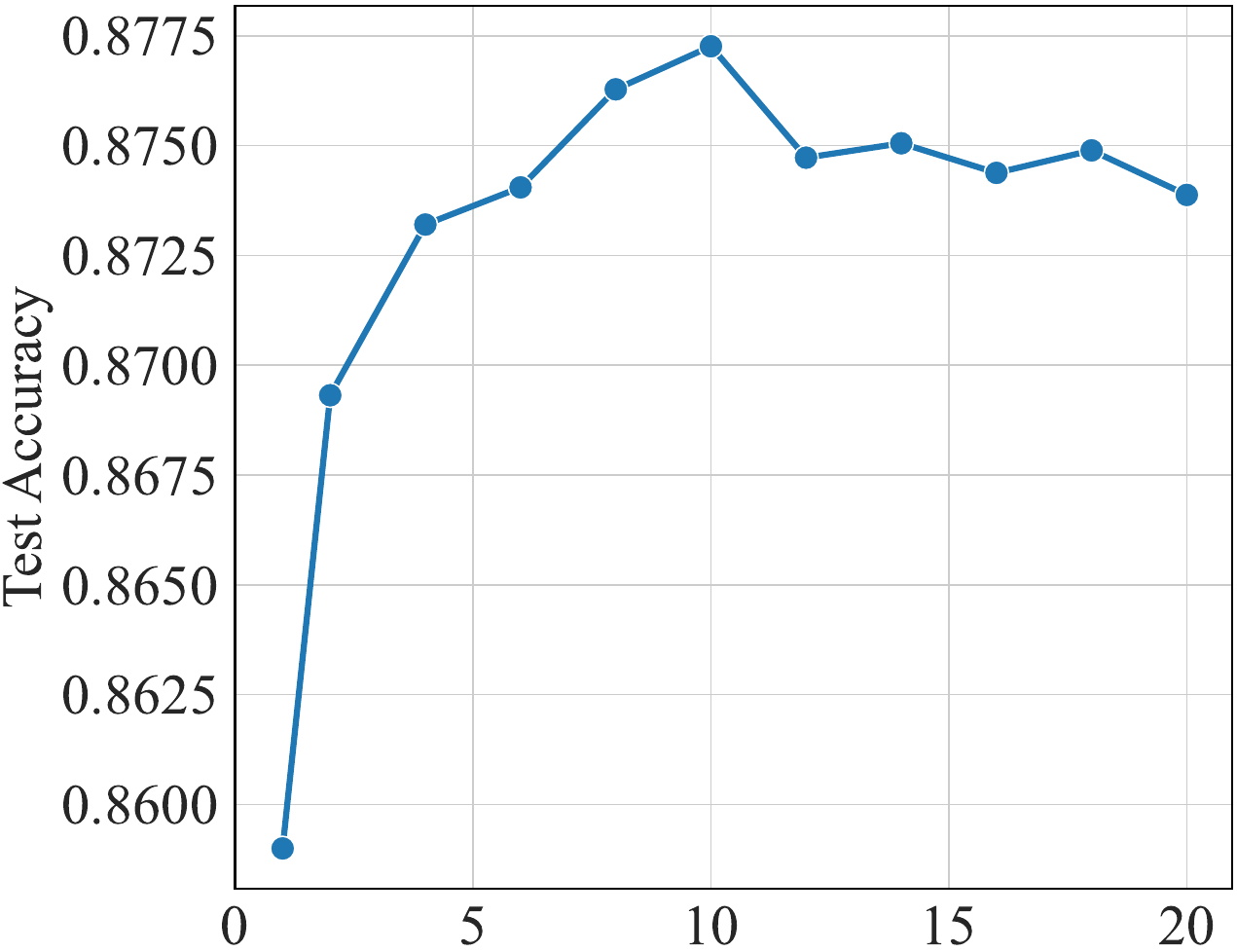}
		\subcaption{Varied updating intervals}
		\label{fig:update_interval}
	\end{minipage}
	\caption{Parameter Sensitivity}
	\label{fig:parameter_sensitivity}
\end{figure}

\subsection{Parameter Sensitivity Analysis (RQ3):}
The prototype-guided cluster discovery mechanism introduces two critical hyperparameters: the number of semantic prototypes $K$, which governs cluster granularity, and the target distribution update interval $T$, which influences training stability. To investigate their effects, we conduct a parameter sensitivity analysis, with results visualized in Figure~\ref{fig:parameter_sensitivity}. 

As shown in Figure~\ref{fig:cluster_num}, the model exhibits a non-monotonic relationship between performance and the number of prototypes on MGTAB. Initially, performance improves as $K$ increases, suggesting that additional prototypes assist the model in capturing diverse user groups. However, beyond an optimal threshold, excessive prototypes diminish detection capability, likely due to overfitting to local features, which reduces the model's generalization ability. 

Figure~\ref{fig:update_interval} illustrates the accuracy under varying update intervals for the target distribution on Twibot-20. The results indicate that updating the target distribution at very short intervals destabilizes model training and leads to significantly poor performance. In contrast, excessively large update intervals may lead to an outdated target distribution, resulting in inefficient model training. Thus, selecting an appropriate update interval is crucial for maintaining stable and efficient training.

\begin{figure}[t]
    \centering
    \includegraphics[width=0.75\linewidth]{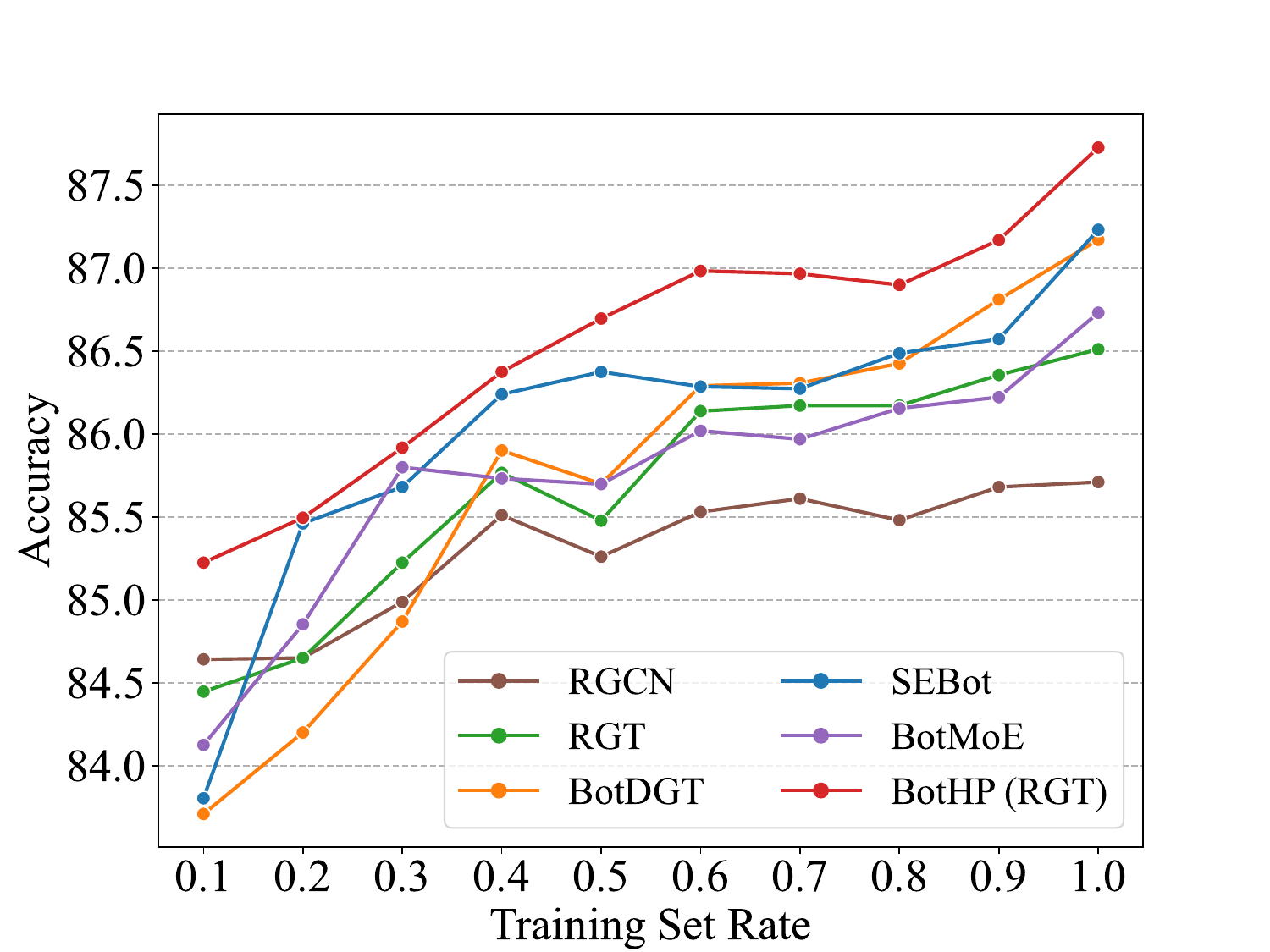}
    \caption{Detection performance with a limited training set ranging from 10\% to 100\%. }
    \label{fig:label}
\end{figure}

\begin{figure*}[t]
    \centering
    \includegraphics[width=0.92\linewidth]{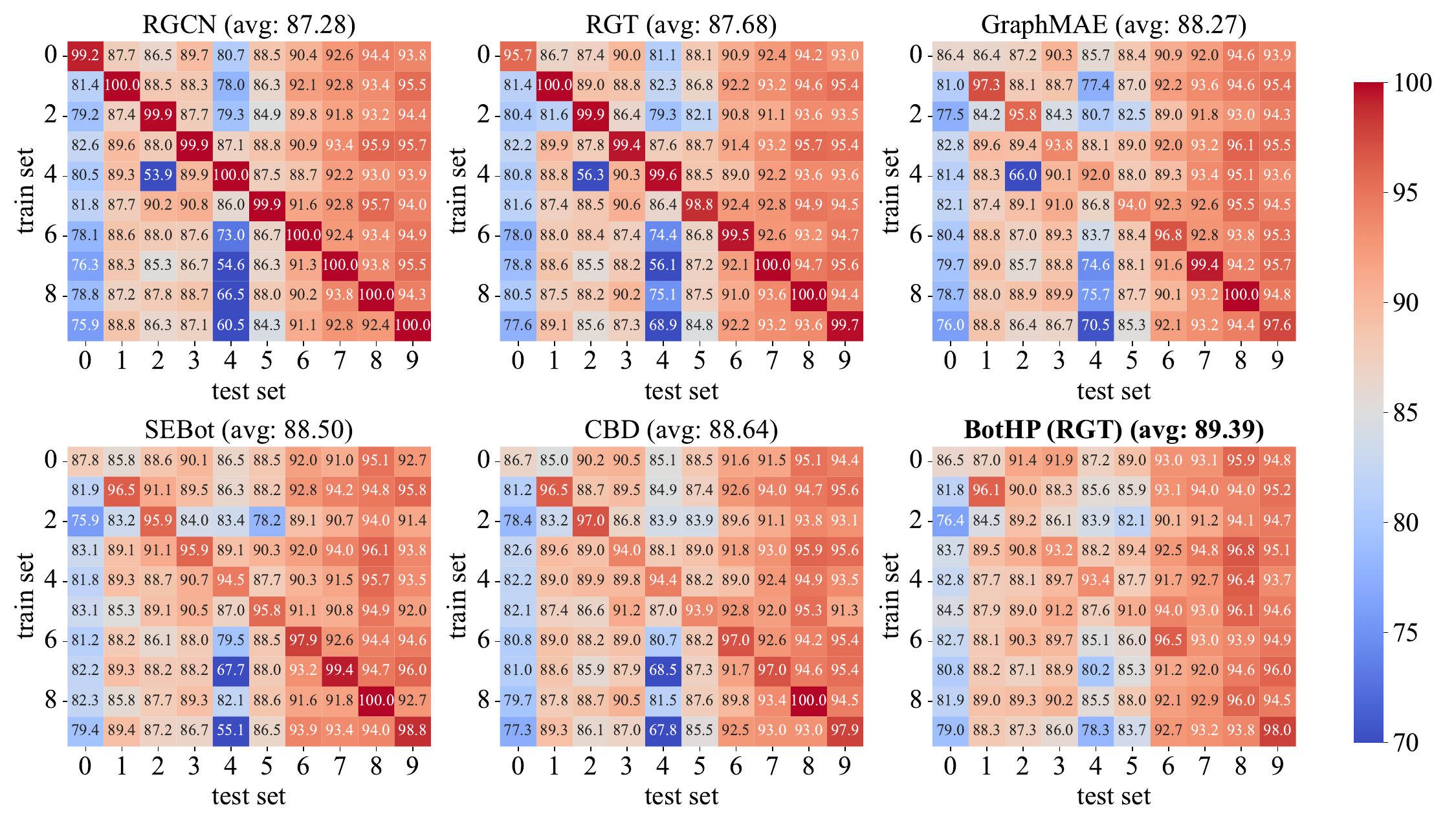}
    \caption{Accuracy scores of different methods trained on the row-indexed community and tested on the column-indexed one. We report the average accuracy across all off-diagonal elements, which serves as a metric for generalization capability.}
    \label{fig:cross}
\end{figure*}

\subsection{Label Efficiency Study (RQ4)}
Given the scarcity of labeled data in bot detection, enhancing the label efficiency of bot detectors with limited annotations is crucial. We conduct a label efficiency experiment to assess the performance of various bot detectors when only a limited amount of labeled data is available. Specifically, we create ablated settings with partial training sets, re-evaluate the bot detection models, and present the performance in Figure~\ref{fig:label}. 

The results indicate that BotHP consistently outperforms all baseline models across all training set sizes, demonstrating superior label efficiency. Notably, BotHP achieves competitive detection performance with its backbone graph-aware encoder, RGT, using only 40\% of the labeled data, which indicates that BotHP effectively alleviates the label reliance of graph-based bot detectors. Furthermore, the performance curves of most supervised detectors exhibit significant fluctuations, suggesting that increasing labeled data does not always lead to performance improvement. We attribute this to a distribution shift between the additional training data and the test set. As supervised detectors tend to overfit the additional training set, they struggle to generalize to the test set. In contrast, BotHP demonstrates greater stability, as its self-supervised learning scheme enables it to learn more robust representations. Further details on the generalization study are provided in Section~\ref{sec:Generalization Study}.

\subsection{Generalization Study (RQ5)}
\label{sec:Generalization Study}

The behavioral characteristics and attack targets of social network bots vary significantly across different communities, exhibiting inherent variance and heterogeneity. Existing bot detectors exhibit limited generalization, as most rely solely on the supervision of training data, leading to overfitting to specific data distributions~\cite{hays2023simplistic,yang2020scalable}.

To investigate the generalization capabilities of different methods, we utilize the Louvain algorithm~\cite{blondel2008fast}, a well-known community detection technique, to partition the MGTAB dataset into 10 sub-communities. In alignment with prior studies~\cite{liu2023botmoe,feng2022twibot}, we treat these sub-communities as distinct data folds and assess the performance of BotHP and various baseline methods by training or fine-tuning them on the $i$-th fold and testing them on the $j$-th fold. Notably, GSL methods undergo pre-training across all communities without label supervision before the fine-tuning phase. From Figure~\ref{fig:cross}, we can draw the following observations:

\begin{itemize}[leftmargin=*]
    \item Supervised graph-based bot detection methods exhibit limited cross-community generalization. While they achieve high accuracy on training communities, their performance significantly declines when applied to new communities. This limitation arises because supervised models tend to overfit to specific patterns in the training data, hindering their ability to generalize to unseen communities.
    \item GSL methods enhance the generalization capability of graph-based detectors. This enhancement stems from leveraging pretext tasks to uncover universal patterns across diverse data distributions, thereby avoiding overfitting to single-community features and improving cross-community bot detection performance.
    \item BotHP demonstrates the strongest generalization capability. First, it achieves a 1.71\% improvement in average accuracy over its graph-aware backbone, highlighting its ability to enhance generalization effectively. This improvement suggests that the dual-encoder architecture, coupled with multiple pretext tasks, plays a crucial role in strengthening the generalization capability of graph-based bot detectors. Additionally, BotHP outperforms other graph self-supervised methods, as its innovative pretraining tasks and encoder architecture facilitate more efficient learning of cross-community patterns, offering a more robust solution for bot detection across diverse communities.
\end{itemize}

\section{Conclusion}
In this paper, we propose BotHP, a generative graph self-supervised learning framework that significantly boosts graph-based bot detectors through two core refinements: heterophily-aware representation learning and prototype-guided cluster discovery. BotHP leverages a dual encoder architecture that simultaneously models homophilic and heterophilic patterns to effectively counter the challenge of interaction camouflage. Furthermore, BotHP incorporates a prototype-guided cluster discovery pretext task to model the latent global consistency of bot clusters and identify spatially dispersed yet semantically aligned bot collectives. Extensive experiments demonstrate the effectiveness of BotHP in improving detection performance, alleviating label reliance, and enhancing generalization capability.
\begin{acks}
This work was supported by the Provincial Key Research and Development Program of Anhui (No. 202423l10050033) and the National Key Research and Development Program of China (No. 2022YFB3105405).
\end{acks}

\bibliographystyle{ACM-Reference-Format}
\balance
\bibliography{kdd}

\clearpage
\appendix
\section{Appendix}
\subsection{Algorithm}
\label{sec:algorithm}
To better understand the proposed framework, we provide the pre-training and fine-tuning pseudocode for BotHP in Algorithm~\ref{al:pretraining} and Algorithm~\ref{al:finetuning}, respectively.

\begin{algorithm}[ht]
    \caption{Pre-training algorithm of BotHP}
    \label{al:pretraining}
    \SetKwInOut{Input}{Input}
    \SetKwInOut{Output}{Output}
    \SetAlgoLined
    \Input{An unlabeled social network graph $\mathcal{G}$, update interval $T$}
    \Output{Pre-trained dual encoder $E_{g}$ and $E_{l}$}
    Randomly initialize the encoders $E_{g}$, $E_{l}$ and the decoders $D_{g}$, $D_{l}$\;

  \For{epoch $t \leftarrow 1, \ldots, epochs$} {
        Generate masked feature matrix $\widetilde{X} \leftarrow$ Eq. (\ref{eq:feature_mask})\;
        Reconstruct features by neighbors $\widetilde{H}^{g}, Z^{g} \leftarrow$ Eq. (\ref{eq:gnn_encoder}-\ref{eq:gnn_decoder})\;
        Calculate loss $\mathcal{L}_{N} \leftarrow$ Eq. (\ref{eq:L_N})\;
        Generate corrupted feature matrix $\overline {X} \leftarrow$ Eq. (\ref{eq:feature_drop})\;
        Reconstruct features by ego node $\overline{H}^{l}, Z^{l} \leftarrow$ Eq. (\ref{eq:mlp_encoder_and_decoder})\;
        Calculate loss $\mathcal{L}_{E} \leftarrow$ Eq. (\ref{eq:L_E})\;
        Calculate user embeddings $H^{g} \leftarrow E_g(X, \mathcal{E})$ and $H^{l} \leftarrow E_l(X)$\;
        Calculate loss $\mathcal{L}_{S} \leftarrow$ Eq. (\ref{eq:L_S})\;
        Compute soft assignment $Q \leftarrow$ Eq.(\ref{eq:Q})\;

        \If{$(t - 1) \mod T == 0$}{
        Update target distribution $P\leftarrow$ Eq. 
        (\ref{eq:P})\;}

        Calculate loss $\mathcal{L}_{C} \leftarrow$ Eq. (\ref{eq:L_D})\;
        Calculate pre-training loss $\mathcal{L}_{P} \leftarrow$ Eq. (\ref{eq:pretrain_loss})\;
        Perform backpropagation on loss $\mathcal{L}_{P}$\;
    }
    \Return dual encoder $E_{g}$ and $E_{l}$
\end{algorithm}

\begin{algorithm}[ht]
    \caption{Fine-tuning algorithm of BotHP}
    \label{al:finetuning}
    \SetKwInOut{Input}{Input}
    \SetKwInOut{Output}{Output}
    \SetAlgoLined
    \Input{A partially labeled social network graph $\mathcal{G}$, the pre-trained encoders $E_{g}$ and $E_{l}$}
    \Output{Optimized $E_{g}$, $E_{l}$, and $D_{d}$ for bot detection}
    Randomly initialize decoder $D_{d}$\;
    
   \While{Not converged}{
        Obtain user representation $H^{g}, H^{l}, U \leftarrow$ Eq. (\ref{eq:user_rep})\;
        Calculate fine-tuning loss $\mathcal{L}_{F} \leftarrow$ Eq. (\ref{eq:finetune_loss})\;
        Perform backpropagation on loss $\mathcal{L}_{F}$\;
    }
    \Return $E_{g}$, $E_{l}$, and $D_{d}$
\end{algorithm}

\subsection{Impact of Dual Encoder Architecture}
\label{sec:Impact of Dual Encoder Architecture}

To investigate the impact of the dual encoder architecture in capturing node commonality while preserving node uniqueness, we conducted a comprehensive analysis combining visual representation and statistical validation. The combined visual and statistical results highlight the dual encoder architecture's complementary strengths: it simultaneously models homophilic and heterophilic patterns, explaining its robust performance across social networks with varying degrees of homophily and heterophily.

\begin{figure}[h]
    \centering
    \includegraphics[width=\linewidth]{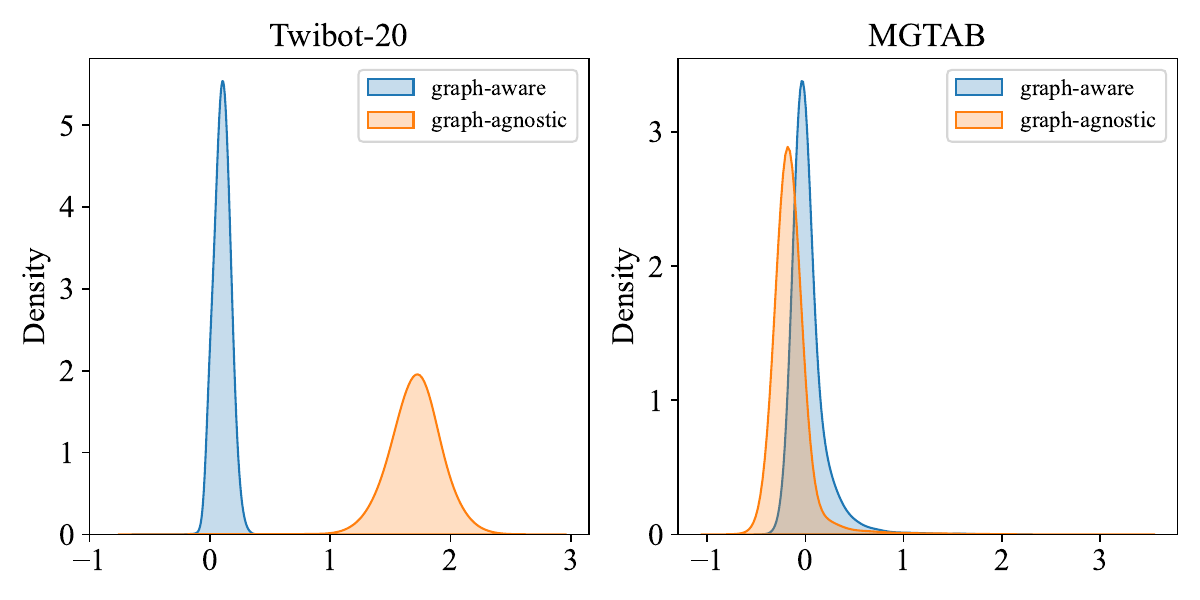}
    \caption{Embedding distribution of Twibot-20 and MGTAB in the first dimension.}
    \label{fig:rep}
\end{figure}

Figure~\ref{fig:rep} illustrates the embedding distribution in the first dimension generated by the graph-aware encoder and the graph-agnostic encoder across the two datasets. Notably, Twibot-20 exhibits high heterophily, whereas MGTAB shows high homophily, as indicated in Table~\ref{tab:dataset}. Our visual analysis reveals that the graph-agnostic encoder consistently produces embeddings with higher standard deviation compared to the graph-aware encoder across both datasets. This observation suggests that the graph-aware encoder tends to reinforce node similarities through neighborhood smoothing. In contrast, the graph-agnostic encoder tends to capture node distinctions, thereby effectively preserving node uniqueness and keeping nodes distinguishable. The discrepancy in standard deviation between the two encoders is particularly pronounced in the heterophilic TwiBot-20 dataset, which is consistent with the heterophily and homophily characteristics of the two datasets.

For statistical validation, we conducted the Wilcoxon Signed-Rank Test to demonstrate that the standard deviation of the embeddings generated by the graph-agnostic encoder is significantly larger than that of the graph-aware encoder. Specifically, representing the embeddings from graph-aware and graph-agnostic encoders as $E^g \in \mathbb{R}^{N \times d_h}$ and $E^l \in \mathbb{R}^{N \times d_h}$ respectively, we calculated the standard deviations along each dimension, denoted as $\sigma^g \in \mathbb{R}^{d_h}$ and $\sigma^l \in \mathbb{R}^{d_h}$. Given the non-normal distribution of these standard deviations and the necessity to evaluate pair-wise differences between $\sigma^g$ and $\sigma^l$ across each dimension, we applied the Wilcoxon Signed Rank Test, which is a non-parametric statistical test suitable for comparing paired samples and without assuming normality of the data. For both Twibot-20 and MGTAB, the p-values were below the significance threshold ($\alpha = 0.05$), indicating that $\sigma^l$ is significantly larger than $\sigma^g$. This quantitative evidence confirms the effect of the proposed dual encoder architecture.

\subsection{Complexity Analysis}
Given a social graph $\mathcal{G} = (\mathcal{V}, \mathcal{E})$, \( n = |\mathcal{V}| \), \( m = |\mathcal{E}|\), and $k$ denotes the number of prototypes, the overall time complexity of BotHP is $O(m + nk)$. The complexity scales linearly with the number of nodes and edges, demonstrating the computational efficiency of BotHP.

\subsection{Representation Learning Study}
\begin{figure}[t]
    \centering
    \subfloat[GCN]{
    \centering
    \includegraphics[width=0.15\textwidth]{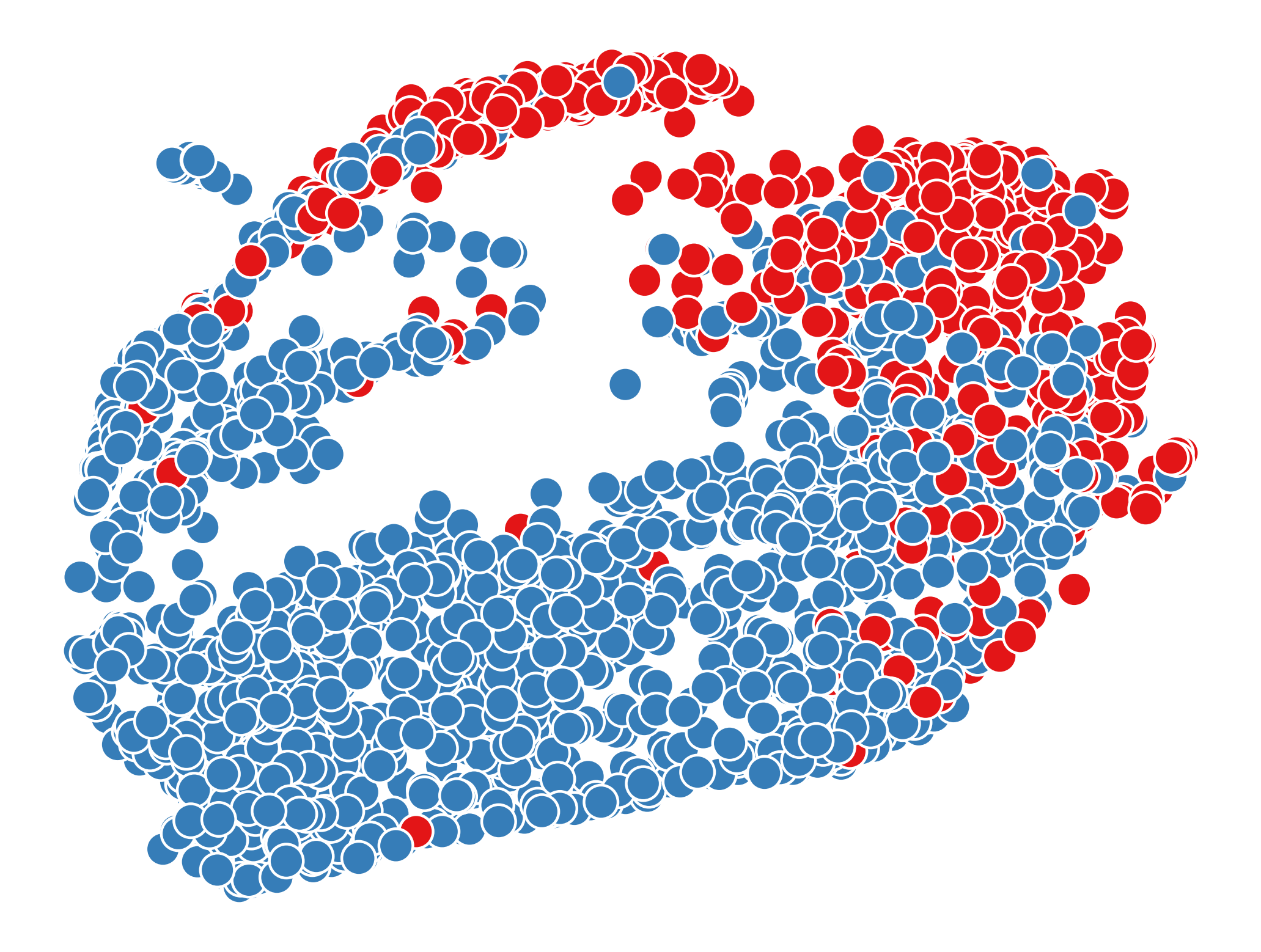}
    }
    \subfloat[FAGCN]{
    \centering
    \includegraphics[width=0.15\textwidth]{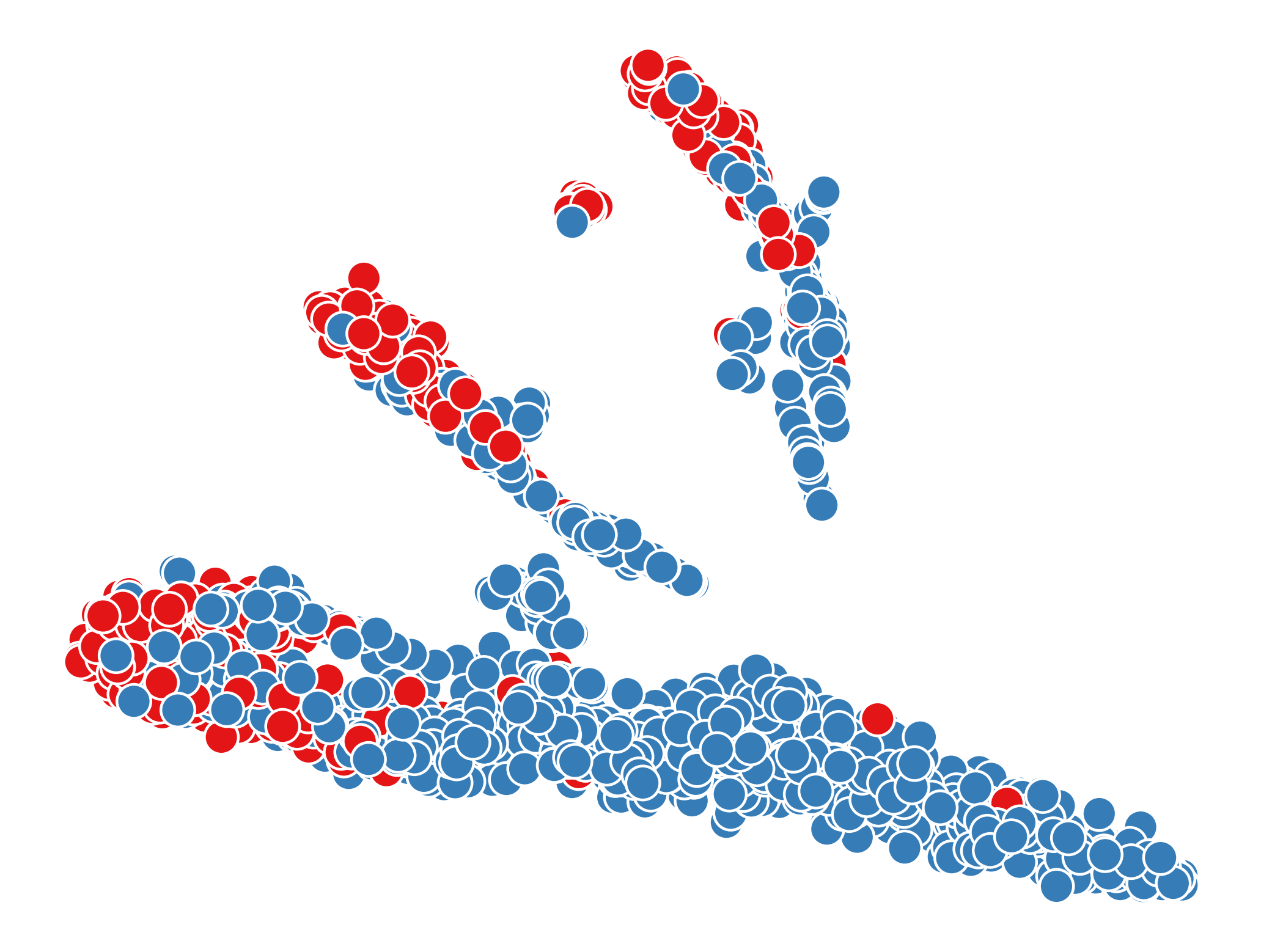}
    }
    \subfloat[RGT]{
    \centering
    \includegraphics[width=0.15\textwidth]{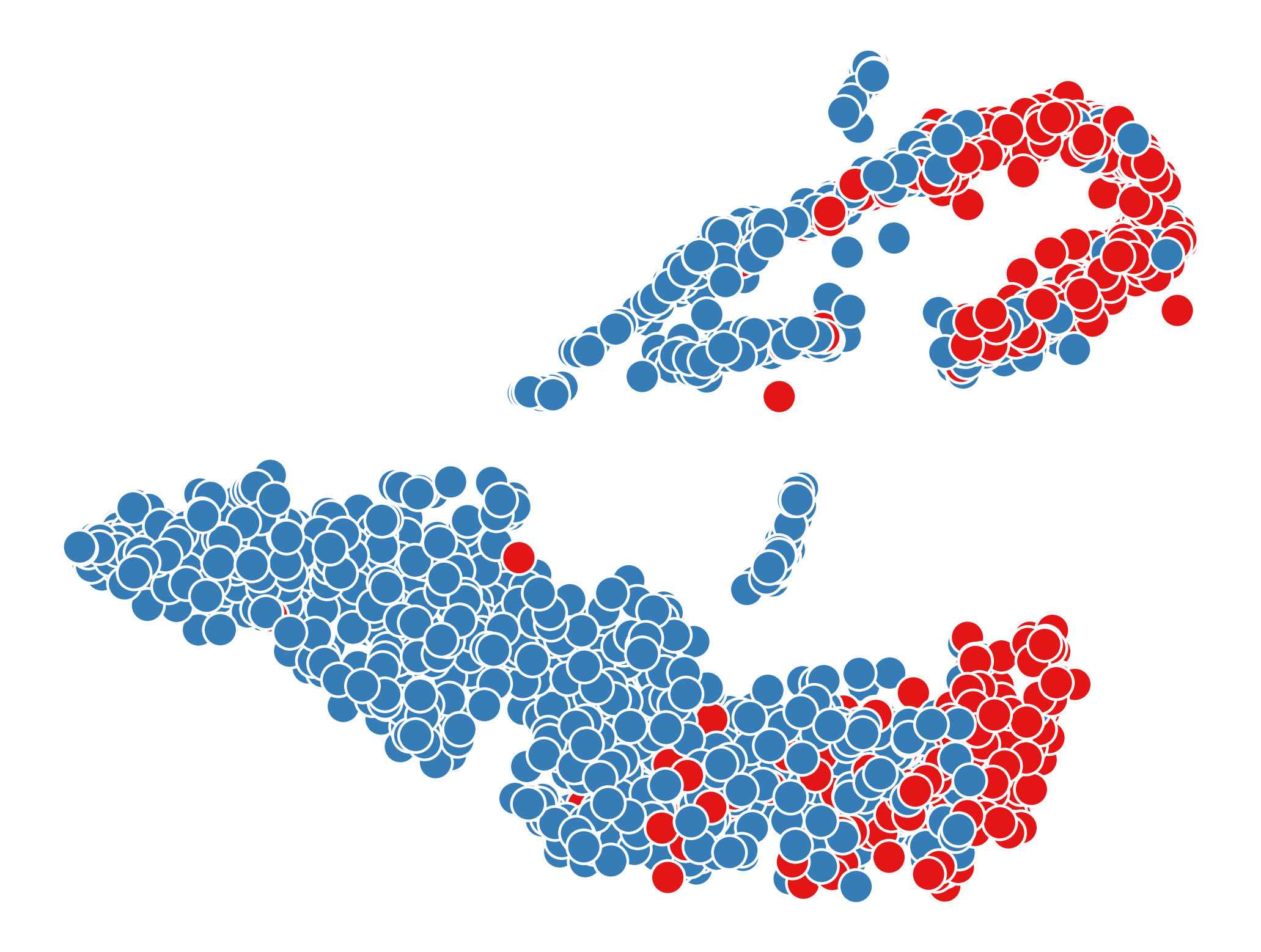}
    }
    \\
    \subfloat[GraphMAE]{
    \centering
    \includegraphics[width=0.15\textwidth]{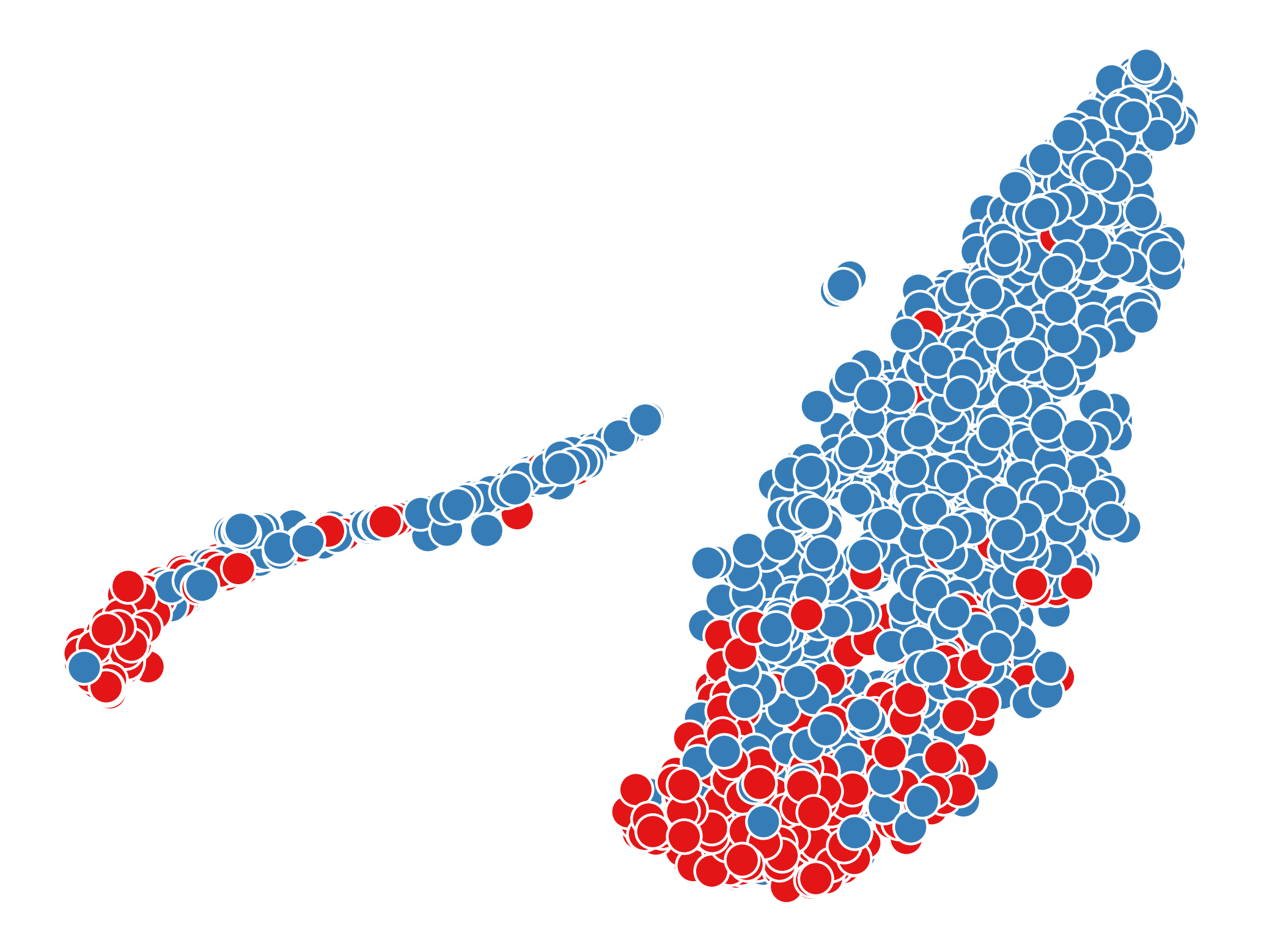}
    }
    \subfloat[SEBot]{
    \centering
    \includegraphics[width=0.15\textwidth]{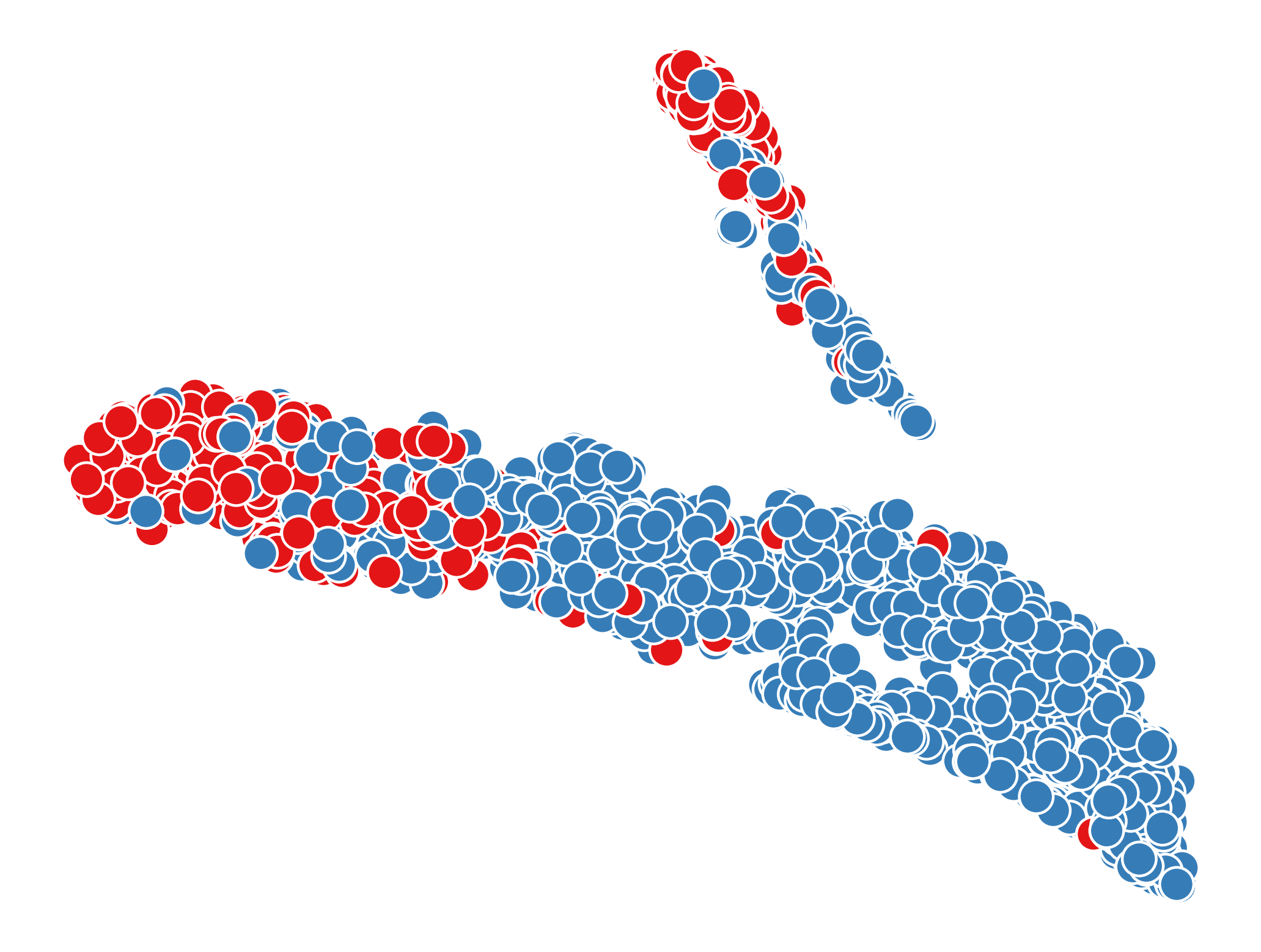}
    }
    \subfloat[BotHP]{
    \centering
    \includegraphics[width=0.15\textwidth]{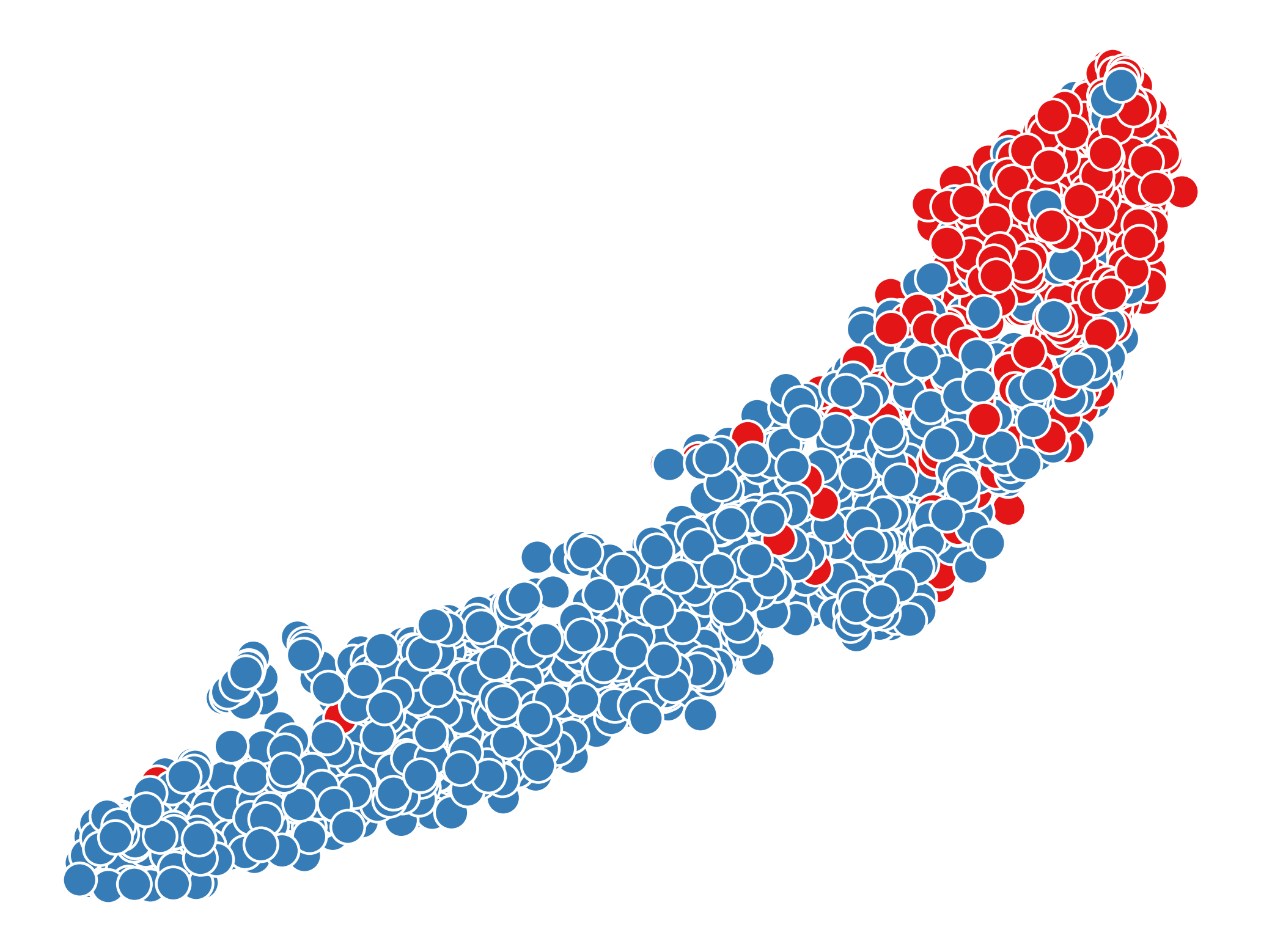}
    }
    \centering
    \caption{User representations visualization on MGTAB. Red represents bots, while blue represents humans.}
    \label{fig:visual} 
\end{figure}

To evaluate the representation learning capability of our proposed framework, we visualize the user representations obtained from different methods using t-SNE~\cite{van2008visualizing} and present the results in Figure~\ref{fig:visual}. For self-supervised learning methods, we present the user representations after fine-tuning on the downstream bot detection task. Compared to other methods, the representations generated by BotHP exhibit discernible clustering, with bots and humans forming separate clusters with minimal overlap. This indicates the discriminative power of BotHP across bots and humans. Furthermore, BotHP generates a single cluster for users within the same category, instead of splitting them into numerous chunks exhibited by other methods. This suggests that BotHP better captures the intrinsic characteristics within each category and preserves intra-class consistency. Overall, BotHP learns more distinctive and comprehensive user representations, making it more effective for bot detection tasks.

\end{document}